\documentclass[longbibliography,final,aps,prx,amsmath,amssymb,
			superscriptaddress]{revtex4-2}
			
\usepackage[T1]{fontenc}
\usepackage{lmodern}
\usepackage{blindtext}
\usepackage{amsfonts}
\usepackage{amsmath}
\usepackage{microtype}
\usepackage[usenames,dvipsnames,table,x11names]{xcolor}
\definecolor{shadecolor}{gray}{0.9}
\usepackage{graphicx}
\usepackage{multirow}
\usepackage{overpic}
\usepackage{booktabs}
\usepackage{tabularx}
\usepackage{algorithm}
\usepackage{algpseudocode}
\usepackage{natbib}
\usepackage[colorlinks,linktoc=all]{hyperref}
\usepackage[all]{hypcap}
\definecolor{darkblue}{rgb}{0.0, 0.0, 0.55}
\definecolor{darkmidnightblue}{rgb}{0.0, 0.2, 0.4}
\definecolor{dukeblue}{rgb}{0.0, 0.0, 0.61}
\definecolor{zaffre}{rgb}{0.0, 0.08, 0.66}
\hypersetup{citecolor=zaffre}
\hypersetup{linkcolor=zaffre}
\hypersetup{urlcolor=zaffre}
\usepackage[capitalize]{cleveref}
\usepackage{mathtools}
\usepackage{layouts}
\usepackage{enumerate}
\usepackage[shortlabels]{enumitem}
\usepackage{mathrsfs}
\usepackage{xparse}
\usepackage[table]{xcolor}
\usepackage{colortbl}
\usepackage{pgfplots}
\usepackage{pgfplotstable}

\pgfplotstableset{
    /color cells/min/.initial=0,
    /color cells/max/.initial=1000,
    /color cells/textcolor/.initial=,
    color cells/.code={%
        \pgfqkeys{/color cells}{#1}%
        \pgfkeysalso{%
            postproc cell content/.code={%
                \begingroup
                \pgfkeysgetvalue{/pgfplots/table/@preprocessed cell content}\value
                \ifx\value\empty
                \endgroup
                \else
                \pgfmathfloatparsenumber{\value}%
                \pgfmathfloattofixed{\pgfmathresult}%
                \let\value=\pgfmathresult
                \pgfplotscolormapaccess
                [\pgfkeysvalueof{/color cells/min}:\pgfkeysvalueof{/color cells/max}]%
                {\value}%
                {\pgfkeysvalueof{/pgfplots/colormap name}}%
                \pgfkeysgetvalue{/pgfplots/table/@cell content}\typesetvalue
                \pgfkeysgetvalue{/color cells/textcolor}\textcolorvalue
                \toks0=\expandafter{\typesetvalue}%
                \xdef\temp{%
                    \noexpand\pgfkeysalso{%
                        @cell content={%
                            \noexpand\cellcolor[rgb]{\pgfmathresult}%
                            \noexpand\definecolor{mapped color}{rgb}{\pgfmathresult}%
                            \ifx\textcolorvalue\empty
                            \else
                            \noexpand\color{\textcolorvalue}%
                            \fi
                            \the\toks0 %
                        }%
                    }%
                }%
                \endgroup
                \temp
                \fi
            }%
        }%
    }
}

\newcommand{\be}{\begin{equation}}
\newcommand{\ee}{\end{equation}}
\newcommand{\bea}{\begin{eqnarray}}
\newcommand{\eea}{\end{eqnarray}}

\newcommand{\f}[2]{\frac{#1}{#2}}
\newcommand{\ccup}[1]{\left\{#1\right\}}
\newcommand{\bup}[1]{\left(#1\right)}

\newcommand{\Inode}{i}
\newcommand{\Inodetwo}{j}

\newcommand{\Inmass}{g}

\newcommand{\Outmass}{h}

\newcommand{\Flux}{P}

\definecolor{darkblue}{rgb}{0, 0.24, 0.64}
\definecolor{darkorange}{rgb}{1, 0.45, 0.0}
\definecolor{darkgreen}{rgb}{0.0, 0.35, 0.0}

\makeatletter
\providecommand*{\cupdot}{%
  \mathbin{%
    \mathpalette\@cupdot{}%
  }%
}
\newcommand*{\@cupdot}[2]{%
  \ooalign{%
    $\m@th#1\cup$\cr
    \hidewidth$\m@th#1\cdot$\hidewidth
  }%
}
\makeatother

\crefformat{equation}{Eq.~#2(#1)#3}
\crefrangeformat{equation}{eqs. #3(#1)#4-#5(#2)#6}
\crefname{eqnarray}{eq.}{eqs.}
\Crefname{eqnarray}{Eq.}{Eq.}
\crefname{figure}{fig.}{figs.}
\Crefname{figure}{Fig.}{Figs.}

\usepackage[caption=false]{subfig}
\usepackage[textsize=tiny,color=SkyBlue]{todonotes}
\usepackage[normalem]{ulem} 

\makeatletter
\def\clearfmfn{\let\@FMN@list\@empty}    
\makeatother

\makeatletter
\def\maketitle{
\@author@finish
\title@column\titleblock@produce
\suppressfloats[t]}
\makeatother

\begin{document}

\title{Immiscible Color Flows in Optimal Transport Networks for Image Classification}
\author{Alessandro Lonardi}
\thanks{Equal contribution}
\email{alessandro.lonardi@tuebingen.mpg.de}
\affiliation{Max Planck Institute for Intelligent Systems, Cyber Valley, T{\"u}bingen 72076, Germany}
\author{Diego Baptista}
\thanks{Equal contribution}
\email{diego.theuerkauf@tuebingen.mpg.de}
\affiliation{Max Planck Institute for Intelligent Systems, Cyber Valley, T{\"u}bingen 72076, Germany}
\author{Caterina De Bacco}
\email{caterina.debacco@tuebingen.mpg.de}
\affiliation{Max Planck Institute for Intelligent Systems, Cyber Valley, T{\"u}bingen 72076, Germany}

\begin{abstract}
In classification tasks, it is crucial to meaningfully exploit the information contained in data. While much of the work in addressing these tasks is devoted to building complex algorithmic infrastructures to process inputs in a black-box fashion, less is known about how to exploit the various facets of the data, before inputting this into an algorithm. Here, we focus on this latter perspective, by proposing a physics-inspired dynamical system that adapts Optimal Transport principles to effectively leverage color distributions of images. Our dynamics regulates immiscible fluxes of colors traveling on a network built from images. Instead of aggregating colors together, it treats them as different commodities that interact with a shared capacity on edges. The resulting optimal flows can then be fed into standard classifiers to distinguish images in different classes. We show how our method can outperform competing approaches on image classification tasks in datasets where color information matters. 
\end{abstract}

\maketitle

\section{Introduction}

Optimal Transport (OT) is a powerful method for computing the distance between two data distributions. This problem has a cross-disciplinary domain of applications, ranging from logistic and route optimization \cite{kaiser2020discontinuous, lonardi2021multicommodity, lonardi2021infrastructure}, to biology \cite{demetci2020gromov, katifori2010damage} and computer vision \cite{werman1985distance,peleg1989unified, rubner1998metric, rubner2000earth,baptista2020principled}, among others. Within this broad variety of problems, OT is largely utilized in machine learning \cite{COTFNT}, and deployed for solving classification tasks, where the goal is to optimally match discrete distributions that are typically learned from data. Relevant usage examples are also found in multiple fields of physics, as in protein fold recognition \cite{koehl_pre},  stochastic thermodynamics \cite{aurell2011optimal}, designing transportation networks \cite{leite2022revealing,baptista2020network}, routing in multilayer networks \cite{ibrahim2021optimal} or general relativity \cite{mondino2022optimal}. 
A prominent application is  image classification \cite{grauman2004fast, cuturisinkhorn, koehl_prl,thorpe2017transportation, pele2008linear, pele2009fast}, where the goal is to measure the similarity between two images. OT solves this problem by interpreting image pairs as two discrete distributions and then assesses their similarity via the Wasserstein ($W_{1}$) distance (\cite{villaniot}, Definition 6.1) a measure obtained by minimizing the cost needed to transform one distribution into the other.  Using $W_1$ for image classification carries many advantages over other similarity measures between histograms. For example, $W_1$ preserves all properties of a metric \cite{villaniot,rubner2000earth},  it is robust over domain shift for train and test data \cite{pele2008linear},  and it provides meaningful gradients to learn data distributions on non-overlapping domains \cite{arjovsky2017wasserstein}.
Because of these, and several others desirable properties, much research effort has been put in speeding up algorithms to calculate $W_{1}$ \cite{cuturisinkhorn,lin2019efficient,dvurechensky2018computational,koehl_pre, koehl_prl}. However, all these methods overlook the potential of using effectively image colors directly in the OT formulation. As a result, practitioners have access to increasingly efficient algorithms, but that do not necessarily improve accuracy in predictions, as we lack a framework that fully exploits the richness of the input information. 

Colored images, originally encoded into three-dimensional histograms---with one dimension per color channel---are often compressed into lower dimensional data using feature extraction algorithms \cite{pele2009fast,rubner2000earth}. 
Here, we propose a different approach that maps the three distinct color histograms to multicommodity flows transported in a network built using images' pixels. We combine recent  developments of OT  with physics insights of capacitated network models \cite{maritan,  ronellenfitsch2016global, hu2013adaptation, corson2010fluctuations, katifori2010damage, kaiser2020discontinuous} to treat colors as mass of different types that flows through  the edges of a network. 
Different flows are coupled together with a shared conductivity, and minimize a unique cost function. This setup is reminiscent of the distinction between modeling the flow of one substance, e.g. water, and modeling flows of multiple substances that do not mix, e.g. immiscible fluids, that share the same network infrastructure. 
By virtue of this multicommodity treatment we achieve stronger classification performance than state-of-the-art OT-based algorithms in real datasets where color information matters.

\section{Problem formulation}

\subsection{Unicommodity Optimal Transport}\label{sec:uniC}
\label{ssec:unicommodityot}

Given two $m,n$-dimensional probability vectors $g,h$, and a positive-valued ground cost matrix $C$, the goal of a standard---unicommodity---OT problem is to find an Optimal Transport  path $P^\star$  satisfying the conservation constraints $\sum_\Inodetwo \Flux_{\Inode \Inodetwo} = \Inmass_\Inode \, \forall \Inode \, \text{ and } \, \sum_\Inode P_{\Inode \Inodetwo} =  \Outmass_\Inodetwo \, \forall \Inodetwo,$ while minimizing $J(g,h) = \sum_{ij} P_{ij} C_{ij}.$

Entries $P^\star_{ij}$ can be interpreted as the mass transported from $g_i$ to $h_j$ when paying a cost $C_{ij}$, while $J^\star$, i.e. $J$ evaluated at $P^\star$,  encodes the minimum effort needed to transport $g$ to $h$. Notably, if all entries $C_{ij}$ are distances between $i$ and $j$, then $J^\star$ is the $W_1$ distance between $\Inmass$ and $\Outmass$ (see \cite{villaniot} for a standard proof and \cite{rubner2000earth} for derivations focusing on the discrete case).

\subsection{Physics-inspired Multicommodity Optimal Transport}
Interpreting colors as mass traveling along a network built from images' pixels (as we define in details below), unicommodity OT could be used to capture the similarity between grayscale images. However, it may not be ideal for colored images, when color information matters. The limitation of unicommodity OT in \Cref{ssec:unicommodityot} is that it does not fully capture the variety of information contained in different color channels, as it is not able to distinguish them. Motivated by this, we tackle this challenge and move beyond this standard setting by incorporating insights from the dynamics of immiscible flows in physics. Specifically, we treat the different pixels' color channels as mass of different types that do not mix, but rather travel and interact on the same network infrastructure, while optimizing a unique cost function. By assuming capacitated edges with conductivities that are proportional to the amount of mass traveling through an edge, we can define a set of ODEs that regulate fluxes and conductivities. These are optimally distributed along a network to better account for color information, while satisfying physical conservation laws. Similar ideas have been successfully used to route different types of passengers in transportation networks \cite{lonardi2021designing,lonardi2021multicommodity,ibrahim2021optimal}.

Formally, we couple together histograms of $M=3$ color channels, the \emph{commodities}, indexed with $a=1,\dots,M$. We define $g^{a},h^{a}$ as $m,n$-dimensional probability vectors of mass of type $a$. More compactly, we define the matrix $G$ with entries $G_{ia}=g^{a}_{i}$ (resp. $H$ for $h$), each containing the intensity of color channel $a$ in pixel $i$ of the first (resp. second) image. These regulate the sources and sinks of mass in our setting. We then enforce conservation of mass for each commodity index $a$: $\sum_i g^a_i = \sum_j h^a_j$. This ensures that all the color mass in the first image should be accounted for in the second one, and viceversa, and this should be valid for each mass type.

Moreover, we define the set $\Pi(G,H)$, containing $( m \times n \times M )$-dimensional tensors $P$ with entries $P_{ij}^{a}$, being transport paths between $g^a$ and $h^a$. These regulate how fluxes of colors of different types travel along a network.
We enforce interaction between transport paths for different commodities by introducing a \emph{shared cost}:
\be
    \label{eqn:sharedcost}
 J_\Gamma(G,H) = \sum_{ij} || P_{ij} ||_2^\Gamma C_{ij},
 \ee
  where $||P_{ij}||_{2} = ( \sum_a {P_{ij}^a}^2 )^{1/2}$ is the 2-norm of the vector $P_{ij}=\bup{P_{ij}^{1},\dots,P_{ij}^{M}}$ and $0 < \Gamma < 4/3$ is a regularization parameter. We take $\Gamma > 0$ since a negative exponent would favor the proliferation of loops with infinite mass \cite{maritan}. Instead,  we conventionally consider $\Gamma < 4/3$ (see \Cref{ssec:otdyn}) since the cost $J_\Gamma$ exhibits the same convexity properties for any $\Gamma > 1$, i.e., it is strictly convex,  and OT paths do not substantially change with $\Gamma$ in this regime \cite{lonardi2021multicommodity}. We can thus formulate its corresponding multicommodity OT problem as that of finding a tensor $P^\star$ solution of:
\begin{equation}
    \label{eqn:multicom_cost}
    J^\star_\Gamma (G,H) = \min_{P \in \Pi(G,H)} J_\Gamma(G,H).
\end{equation}

Notice that for $M = 1$ and $\Gamma = 1$, we recover the standard unicommodity OT setup. 

The problem in \Cref{eqn:multicom_cost} admits a precise physical interpretation.  In fact, it can be recasted into a constrained minimization problem with objective function being the energy dissipated by the multicommodity flows (Joule's law), and with constant total conductivity. Furthermore, transport paths follow Kirchhoff's law enforcing conservation of mass \cite{lonardi2021designing,lonardi2021multicommodity, bonifaci2022physarum} (see Supplementary Material for a detailed discussion).

Noticeably,  $J_\Gamma$ is a quantity that takes into account all the different mass types, and the OT paths $P^\star$ are found through a unique optimization problem.  We emphasize that this is fundamentally different from solving $M$ independent unicommodity problems, where different types of mass are not coupled together as in our setting, and then combining their optimal costs to estimate images' similarity. Estimating $J^\star_\Gamma(G,H)$ gives directly a quantitative and principled measure of the similarity between two images $G$ and $H$. The lower this cost, the higher the similarity of the two images. While this is valid also for the unicommodity cost in \Cref{sec:uniC}, the difference here is that we account differently for the color information as we distinguish different colors via the $M$-dimensional vector $P_{ij}$. The cost in \Cref{eqn:multicom_cost} then properly couples colors by following physical laws regulating immiscible flows. The idea is that, if this information matters for the given classification task,  incorporating it into the minimization problem would output a cost that helps to distinguish images better, e.g. with higher accuracy. 

\section{Methods}

\subsection{Optimal Transport network on images}
\label{ssec:otimages}

\begin{figure}[t]
    \centering
    \includegraphics[width=.8\linewidth]{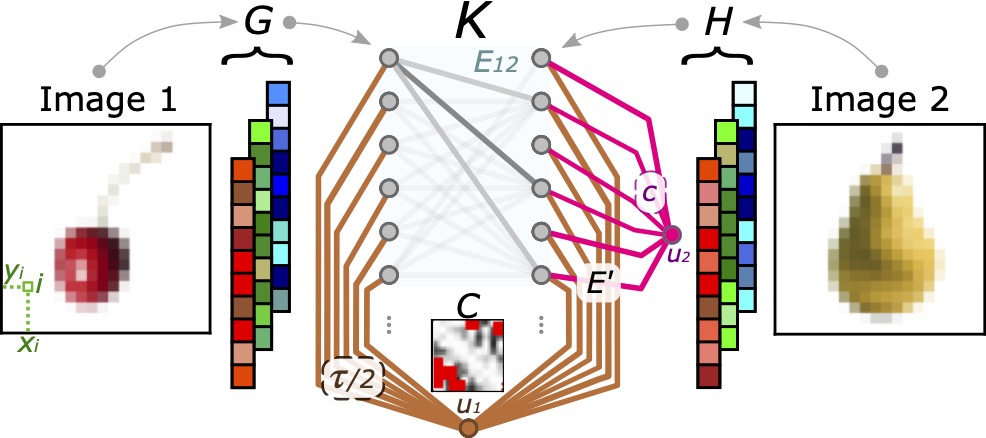}
    \caption{Bipartite network representation for multicommodity OT. The two images (shown in the leftmost and rightmost sides of the panel) are encoded in the RGB matrices $G$ and $H$, that regulate the flow traveling on network $K$. The graph is made of $m+n+2$ nodes, i.e. the total number of pixels plus the two auxiliary vertices introduced in \Cref{ssec:otimages}. Grey edges (belonging to the set $E_{12}$) connect nodes in Image 1 to nodes in Image 2; these edges are trimmed according to a threshold $\tau$. We highlight the entries of the matrix $C$ in red if these are larger than $\tau$. Transshipment and auxiliary edges used to relax mass conservation (which belong to $E'$) are in brown and magenta.}
    \label{fig:model_construction}
\end{figure}

Having introduced the main ideas and intuitions, we now explain in details how to adapt the OT formalism to images. Specifically, we introduce an auxiliary bipartite network $K_{m,n}(V_{1},V_{2},E_{12})$, being the first building block of the network where the OT problem is solved. A visual representation of this is shown in \Cref{fig:model_construction}. The Images 1 and 2 are represented as matrices ($G$ and $H$) of size $m\times M,$ and $n\times M$ respectively, where $M$ is the number of color channels of the images ($M=3$ in our examples). The sets of nodes $V_{1},V_{2}$ of the network $K_{m,n}$ are the pixels of Image 1 and Image 2, respectively. The set of edges $E_{12}$ contains a subset of all pixels' pairs between the two images, as detailed below.
We consider the cost of an edge $(i,j)$ as
\begin{equation}
\label{eqn:costC}
 C_{ij}(\theta, \tau) = \min \{(1-\theta)\,||\mathrm{v}_i-\mathrm{v}_j||_2 + \theta\, || G_{i} - H_{j}||_1, \tau\},
\end{equation}
where the vector $\mathrm{v}_{i} = (x_i,y_i)$ contains the horizontal and vertical coordinates of pixel $i$ of Image 1 (similarly $\mathrm{v}_{j}$ for Image 2). The quantity $\theta \in [0,1]$ is an hyperparameter that is given in input and can be chosen with cross-validation. It acts as a weight for a convex combination between the Euclidean distance between pixels and the difference in their color intensities, following the intuition in \cite{pele2009fast, rubner2000earth}. When $\theta=0$, the OT path $P^\star$ is that minimizing only the geometrical distance between pixels. Instead, when $\theta=1$, pixels' locations are no longer considered, and transport paths are only weighted by color distributions.
The parameter $\tau$ is introduced following \cite{pele2008linear, pele2009fast} with the scope of removing all edges with cost $C_{ij}(\theta, \tau) = \tau$, i.e. those for which $(1-\theta)\,||\mathrm{v}_i-\mathrm{v}_j||_2 + \theta\, || G_{i} - H_{j}||_1>\tau$. These are substituted by $m+n$ transshipment edges $e \in E'$, each of cost $\tau/2$, connected to one unique auxiliary vertex $u_{1}$. Thresholding the cost decreases significantly the computational complexity of OT, making it linear with the number of nodes $|V_1| + |V_2| + 2 = m+n + 2$ (see Supplementary Material).\\
Furthermore, we  relax conservation of mass by allowing $\sum_i G_{ia} \neq \sum_j H_{ja}$. The excess of mass $m^a= \sum_j H_{ja} - \sum_i G_{ia}$ is assigned to a second auxiliary node, $u_{2}$.  We connect it to the network with $n$ additional transshipment edges $e \in E'$, each penalizing the total cost by $c = \max_{ij}{C_{ij}}/2$. This construction improves classification when the histograms' total mass largely differs \cite{pele2008linear}. Intuitively, this can happen when comparing ``darker'' images against ``brighter'' ones. More precisely, when entries of $g^a$ and $h^a$ are further apart in the RGB color space.

Overall we obtain a network $K$ with nodes $V = V_{1} \cup V_{2}  \cup  \ccup{u_{1},u_{2}}$ and edges $E = E_{12}  \cup  E'$, i.e. the original bipartite graph $K_{m,n}$, together with the auxiliary transshipment links and nodes. Note that in its entirety the system is isolated, i.e. the total mass is conserved. See Supplementary Material for a detailed description of the OT setup. 

Given this auxiliary graph, the OT problem is then solved by injecting in nodes $i \in V_{1}$ the color mass contained in Image 1, as specified by $G$, and extracting it in nodes $j \in V_{2}$ of Image 2, as specified by $H$. This is done by transporting mass using either (i) an edge in $E_{12}$, or (ii) a transshipment one in $E'$. In the following Section we describe how this problem is solved mathematically.\\

\subsection{Optimizing immiscible color flows: the dynamics}
\label{ssec:otdyn}

We solve the OT problem by proposing the following ODEs controlling mass transportation:
\begin{alignat}{2}
\label{eqn:kirchhoff}
 \sum_{j \in \partial i} L_{ij}[x] \phi_j^a &= S_i^a \; &&\forall i \in V, a = 1,\dots, M\\
 \label{eqn:dynamics}
 \f{dx_e}{dt} &= x_e^\beta \f{|| \phi_i - \phi_j||^2_2}{C_e^2} - x_e \quad &&\forall e = (i,j) \in E,
\end{alignat}
which constitute the pivotal equations of our model. Here we introduce the \emph{shared conductivities} $x_e \geq 0$, and define $S_{i}^a = G_{ia} - H_{ia}$, taking values $S_{u_{1}}^a = 0$ and $S_{u_{2}}^a = m^a$ on the auxiliary nodes. With $L_{ij}[x] = \sum_e (x_e/C_e) B_{ie}B_{je}$ we denote the weighted Laplacian of $K$, where $B$ is its signed incidence matrix; $\partial i$ is the neighborhood of node $i$. Lastly, $\phi^a_i$ are scalar potentials acting on nodes, for a given commodity $a$. Least-square solutions of \cref{eqn:kirchhoff} are $\phi_i^a [x] = \sum_j L^\dagger_{ij}[x] S^a_i$, where $\dagger$ denotes the Moore-Penrose inverse.  The critical exponent $0 < \beta < 2$ [$\Gamma = 2(2-\beta)/(3-\beta)$] is an hyperparameter that needs to be chosen before solving \Crefrange{eqn:kirchhoff}{eqn:dynamics}.  Depending on the modelling task, its value can be fixed a priori (e.g.  $\beta = 1$ for the shortest path problem \cite{bonifaci2012physarum}, $\beta \simeq 5/3$ for river networks \cite{rinaldo1993self},  and $\beta \to 2^-$ for the Steiner Tree problem \cite{barabasi1996invasion}), or cross-validated as we do here for image classification. The exponent aggregates paths with a principle of economy of scale if $1 < \beta < 2$. It dilutes them along the network otherwise, with the goal of reducing traffic congestion. This behavior is a direct consequence of the subadditivity of $J_\Gamma$ in \Cref{eqn:multicom_cost} for $\beta > 1$ ($\Gamma < 1$), and respectively of its superadditivity for $\beta < 1$ ($\Gamma > 1$). It has been theoretically discussed and empirically observed, for example, in \cite{santambrogio2007optimal,lonardi2021designing,ibrahim2022sustainable}.
 
The feedback mechanism of \Cref{eqn:dynamics} defines multicommodity fluxes ($P_e^a$) that are admissible for the minimization problem introduced in \Cref{eqn:multicom_cost}. Particularly, for color of type $a$ on edges $e = (i,j)$, we couple potentials ($\phi_i^a$) which are solutions of \Cref{eqn:kirchhoff}, and shared conductivities ($x_e$) to define:
\begin{equation}
\label{eqn:P}
P^a_e(t) = x_e(t)\dfrac{\phi^a_i[x(t)] - \phi^a_j[x(t)]}{C_e} \quad \forall e \in E,  a = 1,\dots,M.
\end{equation}
This also highlights another physical interpretation: by interpreting the $\phi^a_i$ as pressure potentials, the fluxes are seen to arise from a difference in pressure between two nodes, as in hydraulic or electrical networks.
Crucially, this allocation is governed by \emph{one unique conductivity} for all commodities, whose dynamics depends on the $2$-norm over $a$ of differences of potentials, as in \cref{eqn:dynamics}. In analogy with immiscible flows, this ensures that flows of different types share the same infrastructure, and in practice it couples them into a unique optimization problem. 

In the case of only one commodity ($M=1$), variants of this dynamics have been used to model transport optimization in various physical systems \cite{ronellenfitsch2016global, hu2013adaptation, corson2010fluctuations, katifori2010damage, kaiser2020discontinuous}. 

The salient result of our construction is that asymptotic trajectories of \Crefrange{eqn:kirchhoff}{eqn:dynamics} are equivalent to minimizers of \cref{eqn:multicom_cost}, i.e. $\lim_{t \to +\infty} P (t) = P^\star$ (see Supplementary Material for derivations following \cite{lonardi2021designing, bonifaci2022physarum}). Therefore, numerically integrating our dynamics  solves the multicommodity OT problem. In other words, this allows us to estimate the optimal cost in \cref{eqn:multicom_cost}, and to use that to compute similarities between images. A pseudo-code of the algorithmic implementation is shown in \Cref{algo:dynamics_alg}.

\begin{algorithm}[H]
  \caption{Multicommodity dynamics}
  \label{algo:dynamics_alg}
   \begin{algorithmic}[1]
    \State{\textbf{Input:} Image 1 ($G \in \mathbb{R}^{m \times M}$), Image 2 ($H \in \mathbb{R}^{n\times M}$), $0 \leq \theta \leq 1$, $\tau \geq 0$, $0 \leq \beta \leq 2$}
    \State{\textbf{Initialize:} $x(0) = \overline{x} > 0$} \Comment{e.g. $\overline{x}_e \sim U(0,1)$}
    \State{Construct a bipartite network $K_{m,n}$ between $G$ and $H$}  \Comment{complexity $O(m \cdot n)$}
    \State{Assign $C_{ij}(\theta, \tau) = \min \{(1-\theta)\,||\mathrm{v}_i-\mathrm{v}_j||_2 + \theta\, || G_{i} - H_{j}||_1, \tau\}$ to every edge $(i,j)$ in $K_{m,n}$, as in \Cref{eqn:costC}} 
    \State{Remove from $K_{m,n}$ all edges s.t. $C_{ij}>\tau$} \Comment{complexity $O(m+n)$}
    \State{Add $u_1$ to $K_{m,n}$ and it $m+n$ auxiliary links,  each costing $\tau/2$}
    \State{Balance mass: add $u_2$, with inflowing mass $m^a = \sum_i H_{ia} - \sum_j G_{ja}$}
    \While {convergence is False}
        \State{Solve Kirchhoff's law,  \Cref{eqn:kirchhoff} $\to \phi \in \mathbb{R}^{|V| \times M}$}
        \State{Update $x$ with discretization of \Cref{eqn:dynamics}}
       \EndWhile
    \State{Compute $P$ as in \Cref{eqn:P}}
    \State{\textbf{Return:} $J^\star_\Gamma(G,H)$ as in \Cref{eqn:multicom_cost}}
   \end{algorithmic}
\end{algorithm}

\subsection{Computational complexity}
In principle, our multicommodity method has a computational complexity of order $O(M|V|^2)$ for complete transport network topologies, i.e., when edges in the transport network $K$ are assigned to all pixels' pairs. Nonetheless, we decrease substantially this complexity by sparsifying the graph with the trimming procedure of \cite{pele2008linear,pele2009fast}, reducing the complexity to $O(M|V|)$. More details are given in the Supplementary Material. 
Empirically, we observe that running \Crefrange{eqn:kirchhoff}{eqn:dynamics}, most of the entries of $x$ decay to zero after a few steps, producing a progressively sparser weighted Laplacian $L[x]$. This allows for faster computation of the Moore-Penrose inverse $L^\dagger[x]$, and of the least-square potentials $\phi_i^a = \sum_j L_{ij}^\dagger[x]S_j^a$ thereon. An experimental thorough analysis of convergence properties of the OT dynamics has been done in \cite{facca2020fast}.

\section{Results and discussion}

\subsection{Classification task}

We provide empirical evidence that our multicommodity dynamics outperforms competing OT algorithms on classification tasks. As anticipated above, we use the OT optimal cost $J^\star_\Gamma$ as a measure of similarity between two images, and perform supervised classification with a $k$-nearest neighbor ($k$-NN) classifier as in \cite{koehl_prl}.  Alternative methods (e.g. SVM as in  \cite{cuturisinkhorn}) could also be employed for this task. However, these may require the cost $J^\star_\Gamma$ to satisfy the distance axioms to properly induce a kernel.  While it is not straightforward to verify these conditions for the OT cost in \Cref{eqn:multicom_cost}, this is not necessary for the $k$-NN classifier, which  requires  instead looser conditions on $J^\star_\Gamma$.

\begin{table}[b]
    \centering
    \begin{tabular}{llcccccc} \toprule
    & Algorithm & \multicolumn{5}{c}{Hyperparameters} & {Class. accuracy} \\
    &  & $\theta$ & $\tau$ & $\beta$ & $\varepsilon$ & $k$ & [\%] ($\uparrow$) \\
    \midrule
    \multirow{4}{*}{\rotatebox[origin=c]{90}{JF30}}
    &Multicommodity & 0.25 & 0.125 & 1 & --- & 1 & \textbf{62.2} \\
    &Sinkhorn RGB & 0.25 & 0.05 & --- & 100 & 1 & 58.4\\
    &Sinkhorn GS & 0.25 & 0.05 & --- & 500 & 1 & 54.3\\
    &Unicommodity & 0.25 & 0.125 & 1.25 & --- & 1 & 53.6\\
    \midrule
    \multirow{4.0}{*}{\rotatebox[origin=c]{90}{FD}}
    &Multicommodity & 0 & 0.04 & 1.5 & --- & 2 & \textbf{75.0}\\
    &Sinkhorn RGB & 0.5 & 0.06 & --- & 750 & 1 &69.6\\
    &Unicommodity & 0 & 0.06 & 1.5 & --- & 5 &64.3\\
    &Sinkhorn GS & 0.25 & 0.06 & --- & 500 & 4 &60.7\\
    \bottomrule
    \end{tabular}
    \caption{Classification task results. With Multicommodity, Sinkhorn RGB, Unicommodity and Sinkhorn GS we label methods on colored images (first two), and grayscale ones (second two). Optimal parameters in the central columns are selected with a 4-fold cross-validation;  $k$ is the number of nearest neighbors used in the classifier. The rightmost column shows the fraction (in percentage) of correctly classified images. Results are ordered by performance, in bold we highlight the best ones. \label{tab:results}}
\end{table}

We compare classification accuracy of our model against: (i) Sinkhorn algorithm \cite{cuturisinkhorn, flamary2021pot} (utilizing the more stable Sinkhorn scheme proposed in \cite{schmitzer2019stabilized}); (ii) a unicommodity dynamics executed on grayscale images, i.e., with color information compressed in one single commodity ($M=1$); (iii) Sinkhorn algorithm on grayscale images. All methods are tested on two datasets: the Jena Flowers 30 Dataset (JF30) \cite{seeland2017plant}, and the Fruit Dataset (FD) of \cite{alves2018handwritten}. The first consists of 1,479 images of 30 wild-flowering angiosperms (flowers). Flowers are labelled with their species, inferring them is the goal of the classification task. The second dataset contains 15 fruit types and 163 images, here we want to classify fruit types.  Parameters of the OT problem setup ($\theta$ and $\tau$), as well as regularization ones ($\beta$ and $\varepsilon$, which enforces the entropic barrier in Sinkhorn algorithm \cite{cuturisinkhorn}), have been cross-validated for both datasets (see Section III and Section IV in Supplementary Material). All methods are then tested in their optimal configurations (see Supplementary Material for implementation details).

Classification results are shown in \cref{tab:results}. In all cases, leveraging colors leads to higher accuracy (about 8$\%$ increase) with respect to classification performed using grayscale images. This signals that, in the datasets under consideration, color information is a relevant feature for differentiating image samples. Remarkably, we get a similar increase in performance (about 7-8$\%$) on both colored datasets, when comparing our multicommodity dynamics against Sinkhorn. As the two algorithms use the same (colored) input, we can attribute this increment to an effective usage of the color that our approach is capable of.

In addition, analyzing results in more detail, we first observe that on JF30 all methods perform best when $\theta = 0.25$, i.e., $25\%$ percent of the information used to build $C$ comes from colors. This trend does not recur on FD, where both dynamics favor $\theta = 0$ (Euclidean $C$). Hence, our model is able to leverage color information via the multicommodity OT dynamical formulation.

Second, on JF30 both dynamics perform best with $\tau=0.125$, contrarily to Sinkhorn-based methods that prefer $\tau = 0.05$. Thus, Sinkhorn's classification accuracy is negatively affected both by low $\tau$---many edges of the transport network are cut, and by $\tau$ large---noisy color information is used to build $C$. 
We do not observe this behavior in our model,  where trimming less edges is advantageous. All optimal values of $\tau$ are lower on FD, since color distributions of this dataset are naturally light-tailed (See Supplementary Material).

Lastly, we investigate the interplay between $\theta$ and $\beta$. We notice that $\theta = 0$ (FD) corresponds to higher $\beta = 1.5$. Instead, for larger $\theta = 0.25$ (JF30), the model prefers lower $\beta$ ($\beta = 1,1.25$ for the multicommodity and the unicommodity dynamics, respectively). In the former case ($\theta=0$, $C_{ij}$ is the Euclidean distance), the cost is  equal to zero for pixels with the same locations. Thus, consolidation of transport paths---large $\beta$---is favored on cheap links. Instead, increasing $\theta$ leads to more edges with comparable costs as colors distribute smoothly over images. In this second scenario, better performance is achieved with distributed transport paths, i.e. lower $\beta$ (See Supplementary Material).

\subsection{Performance in terms of sensitivity}

We assess the effectiveness of our method against benchmarks by comparing the sensitivity of our multicommodity dynamics and that of Sinkhorn algorithm on the colored JF30 dataset.
Specifically, we set all algorithms' parameters to their best configurations, as shown in \Cref{tab:results}. Then, for each of the 30 class in JF30 we compute its one-vs-all sensitivity, i.e. the true positive rate. This is defined for any class $c$ as
\begin{equation}
\label{eqn:sens}
S(c) = \frac{\textrm{TP}(c)}{\textrm{TP}(c) + \textrm{FN}(c)} \, ,
\end{equation}
where $\textrm{TP}(c)$ is the true positives rate, i.e. the number of images in $c$ that are correctly classified; $\textrm{FN}(c)$ is the false negative rate, i.e. the number of $c$-samples that are assigned a label different than $c$. Hence, \Cref{eqn:sens} returns the probability that a sample is assigned label $c$, given that it belongs to $c$.

\begin{figure}[t]
\centering
\includegraphics[width=0.9\linewidth]{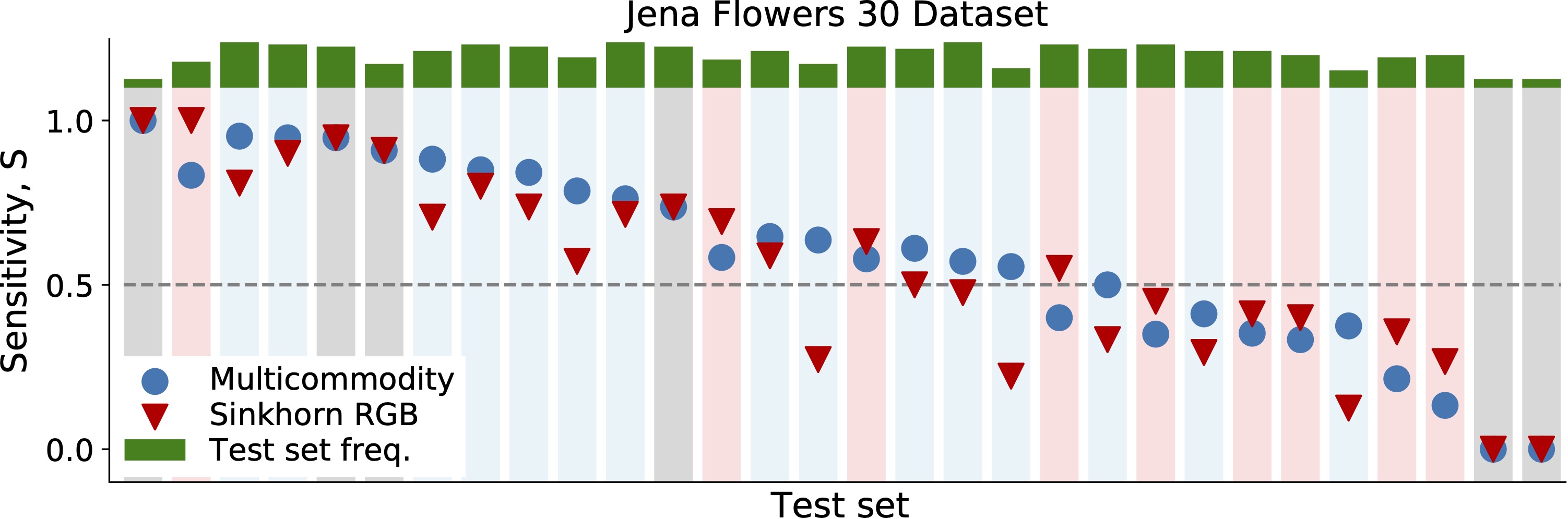}
\caption{Sensitivity on the JF30 dataset.  Sensitivity values are shown for the multicommodity dynamics (blue circles) and for Sinkhorn RGB (red triangles). Markers are sorted in descending order of $S$, regardless the method.  Background colors are blue, red, and gray, when $S$ is higher for the multicommodity method, Sinkhorn, or none of them, respectively. In green we plot frequency bars for all classes in the test set.}
\label{fig:sensitivity}
\end{figure}

We find that our method robustly outperforms Sinkhorn. Specifically,  the multicommodity dynamics has the highest sensitivity $50\%$ of the times---$15$ classes out of the total $30$, as shown in \Cref{fig:sensitivity}. In $9$ cases Sinkhorn has higher sensitivity, and for $6$ classes both methods give the same values of $S$.\\
Furthermore we find that in $2/3$ ($20$ out of $30$) of the classes, the multicommodity dynamics returns $S(c) \geq 1/2$. This means that our model predicts the correct label more than $50 \%$ of the times. In only $3$ out of these $20$ cases Sinkhorn attains higher values of $S$, while in most instances where Sinkhorn outperforms our method it has lower sensitivity $S < 1/2$. Hence, this is the case of classes where both methods have difficulties in distinguishing images.

\subsection{The impact of colors}

To further assess the importance of leveraging color information we conduct three different experiments where we highlight, both qualitatively and quantitatively, various performance differences between the unicommodity and the multicommodity approaches. As the two  share the same principled dynamics based on OT, with the main difference being that multicommodity does not compress the color information, we can use this analysis to better understand how  fully exploiting the color information drives better classification.

\begin{figure*}[t]
\centering
\includegraphics[width=0.8\linewidth]{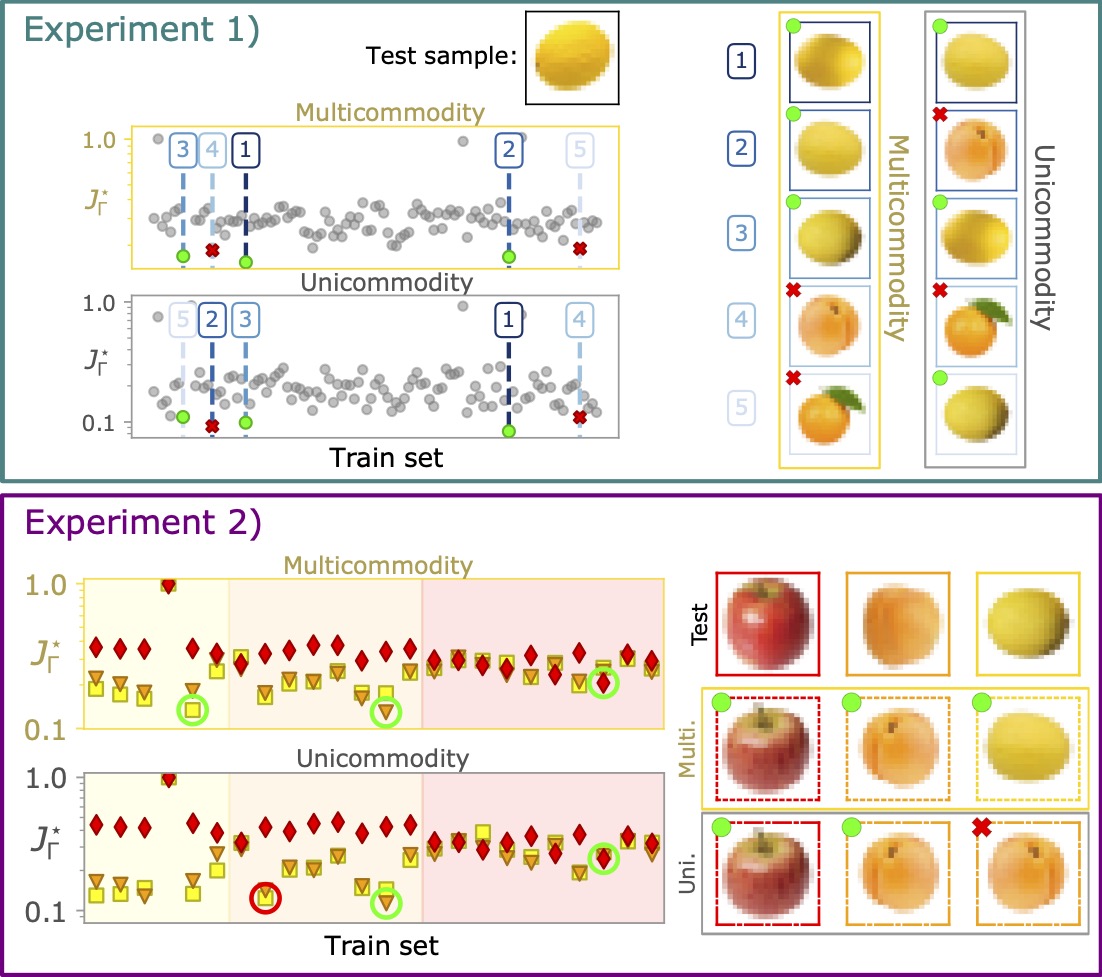}
\caption{Evaluating the effect of colors. Experiment 1: Top black framed image is that to be classified. Predictions given by the multicommodity and the unicommodity dynamics (those with lower $J_\Gamma^\star$) are shown in the right side of the panel, and are displayed in sorted fashion from worst to best (from bottom to top). Experiment 2: Top right samples are the three test images to be classified. Middle and bottom rows are predictions given by the two dynamics. Markers, backgrounds and test images shared color code: red for apples, orange for apricots, yellow for melons. In both panels: green circles and red crosses are used to highlight classified and misclassified images, respectively. All algorithms are executed with their optimal configurations in \cref{tab:results}.}
\label{fig:cp_image_big}
\end{figure*}

\emph{Experiment 1: landscape of optimal cost.} Here we focus on a qualitative comparison between the cost landscapes obtained with the two approaches. We consider the example of an individual image taken from the test set of FD and we plot the landscape of optimal costs $J_\Gamma^\star$ when comparing it against the train set. Results for the multicommodity dynamics ($M=3$), and for the unicommodity dynamics ($M=1$) on grayscale images are in \Cref{fig:cp_image_big}. Here, we highlight the $5$ lowest values of the cost and mark them in green if they correspond to correctly classified train samples, and in red otherwise. While at a first glance one may conclude that their performance is identical (as both dynamics classify correctly $3$ samples out of $5$), we notice how the multicommodity dynamics consistently clusters them at the bottom of the cost landscape, thus ranking them in a better order. This may be explain why the cross-validated best value of $k$ (the number of nearest-neighbors in the $k$-NN classifier) is higher for unicommodity methods in this dataset. On a larger sample of data, this results in better overall classification performance, as show in \Cref{tab:results}.

\emph{Experiment 2: controlling for shape.} We further mark this tendency with a second experiment where we select a subset of FD composed of images belonging to three classes of fruits that have similar shapes but different colors: red apples, orange apricots, yellow melons. As we expect shape to be less informative than colors in this custom set, we can  assess the extent to which color plays a crucial role in the classification process. Specifically, the test set is made of three random samples each drawn from one of these classes (top row of the rightmost panel) in \Cref{fig:cp_image_big}, while the train set contains the remaining instances of the classes. We plot the cost landscape $J_\Gamma^\star$ for the train set and draw in red, orange, and yellow values of $J_\Gamma^\star$ correspondent to samples which are compared against the test apple, apricot, and melon, respectively. We also sort the train samples so that they are grouped in three regions (highlighted by the background color in \Cref{fig:cp_image_big}), correspondent to train melons, apricots, and apples.
With this construction, if the minimum cost among the yellow markers falls in the yellow region, it will correspond to a correctly-classified sample (resp. for orange and red). We further mark the yellow, orange, and red minima in green if test and train labels correspond, i.e. markers' and background's colors are the same, and in red otherwise. Train and test samples are also in \Cref{fig:cp_image_big}.
The multicommodity dynamics  correctly labels each test image. In contrast, the unicommodity dynamics fails at this task, labeling a melon as an apricot. This suggests that the multicommodity approach is able to use the color information in datasets where this feature is more informative than others, e.g. shape.

\begin{figure}[t]
\centering
\includegraphics[width=0.8\linewidth]{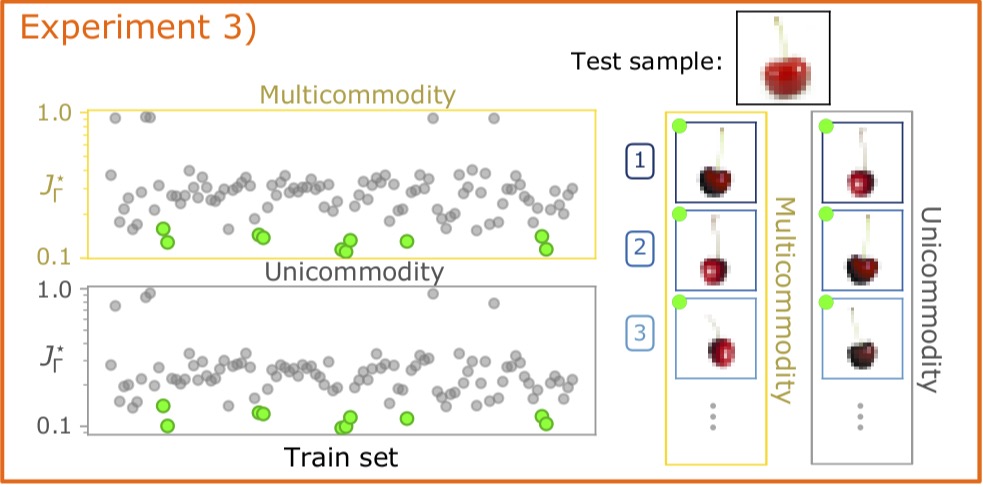}
\caption{Evaluating the importance of colors: when shapes matter most. Experiment 3: Top black framed image is that to be classified. The best 3 (out of 10) predictions returned by the two dynamics are shown on the right.  We mark with green circles training samples belonging to the same class of the test image. All algorithms are executed with their optimal configurations in \cref{tab:results}.}
\label{fig:cp_image_small}
\end{figure}

\emph{Experiment 3: when shape matters.} Having shown results on a custom dataset where shape was controlled to matter less, we now do the opposite and select a dataset where this feature should be more informative. The goal is to assess whether a multicommodity approach helps in this case as well, as its main input information may not be as relevant anymore. Specifically, we select as a test sample a cherry, whose form is arguably distinguishable from that of many other fruits in the dataset. One can expect that comparing it against the train set of FD will result in having \emph{both} unicommodity and multicommodity dynamics able to assign low $J_\Gamma^\star$ to train cherries, and higher costs to other fruits. This intuition is confirmed by results in \Cref{fig:cp_image_small}. Here train cherries (in green) strongly cluster in the lower portion of the cost landscape, whereas all the other fruits have higher costs. In \Cref{fig:cp_image_small} we also plot some of the correctly classified train samples. 
These results suggest that when color information is negligible compared to another type of information (e.g. shape), unicommodity and multicommodity formulations perform similarly. In light of this, we reinforce the claim that our multicommodity formulation can boost classification in contexts where color information does matter, but may not give any advantage when other types of information is more informative. We encourage practitioners to evaluate when this is the case based on domain-knowledge when available.

\section{Conclusion}

We propose a physics-informed multicommodity OT formulation for effectively using color information to improve image classification. We model colors as immiscible flows traveling on a capacitated network, and propose equations for its dynamics, with the goal of optimizing flow distribution on edges.  Color flows are regulated by a shared conductivity, and minimize a unique cost function. Thresholding the ground cost as in \cite{pele2008linear, pele2009fast} makes our model computationally efficient. 

We outperform other OT-based approaches as the Sinkhorn algorithm on two datasets where color matters. Our model also assigns lower cost to correctly classified images than its unicommodity counterpart, and it is more robust on datasets where items have similar shape, and thus color information is distinctly relevant. We note that, for some datasets, color information may not matter, as when another type of information (e.g. shape) has stronger discriminative power. However, while we focused here on different color channels as the different commodities in our formulation, the ideas of this paper can be extended to scenarios where other  relevant information can be distinguished into different types. For instance, one could  combine several features together, e.g. colors, contours and objects' orientations, when available.

Our model can be further improved. While it uses the thresholding of  \cite{pele2008linear,pele2009fast} to speed-up convergence (as mentioned in \Cref{ssec:otimages}), it is still slower than Sinkhorn-based methods. Hence, investigating approaches aiming at improving its computational performance is an important direction for future work. Speed-up can be achieved, for example,  with the implementation of \cite{facca2020fast}, where the unicommodity OT problem on sparse topologies is solved in $O(|E|^{0.36})$ time steps. This bound has been found using a backward Euler scheme combined with the inexact Newton-Raphson method for the update of $x$, and solving Kirchhoff’s law using an algebraic multigrid method \cite{trottenberg2000multigrid}.

Our main goal is to frame an image classification task  into that of finding optimal flows of mass of different types in networks built from images. We follow physics principles to assess whether using colors as immiscible flows can give an advantage compared to other standard OT-based methods that do not incorporate such insights. The increased classification performance observed in our experiments stimulates the integration of similar ideas into deep network architectures  \cite{eisenberger2022unified} as a relevant avenue for future work. Combining their prediction capabilities with our insights on how to better exploit the various facets of the input data has the potential to push even further the performance of deep classifiers. For example, one could extend the state-of-the-art architecture of \citet{eisenberger2022unified},  which efficiently computes implicit gradients for generic Sinkhorn layers within a neural network,  by including edge, shape, and contour information for Wassertein barycenters computation or image clustering.

\section*{Code and data availability statement}
To facilitate practitioners using our algorithms, we make our Python code publicly available \cite{git_repo}. The datasets Jena Flowers 30 Dataset (JF30) and the Fruit Dataset (FD) used for this study can be found in \cite{seeland2017plant} and \cite{alves2018handwritten}, respectively.



\section*{Acknowledgments}
The authors thank the International Max Planck Research School for Intelligent Systems (IMPRS-IS) for supporting Alessandro Lonardi and Diego Baptista.



\bibliography{bibliography}

\begin{thebibliography}{53}%
\makeatletter
\providecommand \@ifxundefined [1]{%
 \@ifx{#1\undefined}
}%
\providecommand \@ifnum [1]{%
 \ifnum #1\expandafter \@firstoftwo
 \else \expandafter \@secondoftwo
 \fi
}%
\providecommand \@ifx [1]{%
 \ifx #1\expandafter \@firstoftwo
 \else \expandafter \@secondoftwo
 \fi
}%
\providecommand \natexlab [1]{#1}%
\providecommand \enquote  [1]{``#1''}%
\providecommand \bibnamefont  [1]{#1}%
\providecommand \bibfnamefont [1]{#1}%
\providecommand \citenamefont [1]{#1}%
\providecommand \href@noop [0]{\@secondoftwo}%
\providecommand \href [0]{\begingroup \@sanitize@url \@href}%
\providecommand \@href[1]{\@@startlink{#1}\@@href}%
\providecommand \@@href[1]{\endgroup#1\@@endlink}%
\providecommand \@sanitize@url [0]{\catcode `\\12\catcode `\$12\catcode
  `\&12\catcode `\#12\catcode `\^12\catcode `\_12\catcode `\%12\relax}%
\providecommand \@@startlink[1]{}%
\providecommand \@@endlink[0]{}%
\providecommand \url  [0]{\begingroup\@sanitize@url \@url }%
\providecommand \@url [1]{\endgroup\@href {#1}{\urlprefix }}%
\providecommand \urlprefix  [0]{URL }%
\providecommand \Eprint [0]{\href }%
\providecommand \doibase [0]{https://doi.org/}%
\providecommand \selectlanguage [0]{\@gobble}%
\providecommand \bibinfo  [0]{\@secondoftwo}%
\providecommand \bibfield  [0]{\@secondoftwo}%
\providecommand \translation [1]{[#1]}%
\providecommand \BibitemOpen [0]{}%
\providecommand \bibitemStop [0]{}%
\providecommand \bibitemNoStop [0]{.\EOS\space}%
\providecommand \EOS [0]{\spacefactor3000\relax}%
\providecommand \BibitemShut  [1]{\csname bibitem#1\endcsname}%
\let\auto@bib@innerbib\@empty
\bibitem [{\citenamefont {Kaiser}\ \emph {et~al.}(2020)\citenamefont {Kaiser},
  \citenamefont {Ronellenfitsch},\ and\ \citenamefont
  {Witthaut}}]{kaiser2020discontinuous}%
  \BibitemOpen
  \bibfield  {author} {\bibinfo {author} {\bibfnamefont {F.}~\bibnamefont
  {Kaiser}}, \bibinfo {author} {\bibfnamefont {H.}~\bibnamefont
  {Ronellenfitsch}},\ and\ \bibinfo {author} {\bibfnamefont {D.}~\bibnamefont
  {Witthaut}},\ }\bibfield  {title} {\bibinfo {title} {Discontinuous transition
  to loop formation in optimal supply networks},\ }\href
  {https://doi.org/10.1038/s41467-020-19567-2} {\bibfield  {journal} {\bibinfo
  {journal} {Nature communications}\ }\textbf {\bibinfo {volume} {11}},\
  \bibinfo {pages} {1} (\bibinfo {year} {2020})}\BibitemShut {NoStop}%
\bibitem [{\citenamefont {Lonardi}\ \emph {et~al.}(2022)\citenamefont
  {Lonardi}, \citenamefont {Putti},\ and\ \citenamefont
  {De~Bacco}}]{lonardi2021multicommodity}%
  \BibitemOpen
  \bibfield  {author} {\bibinfo {author} {\bibfnamefont {A.}~\bibnamefont
  {Lonardi}}, \bibinfo {author} {\bibfnamefont {M.}~\bibnamefont {Putti}},\
  and\ \bibinfo {author} {\bibfnamefont {C.}~\bibnamefont {De~Bacco}},\
  }\bibfield  {title} {\bibinfo {title} {Multicommodity routing optimization
  for engineering networks},\ }\href
  {https://doi.org/10.1038/s41598-022-11348-9} {\bibfield  {journal} {\bibinfo
  {journal} {Scientific Reports}\ }\textbf {\bibinfo {volume} {12}},\ \bibinfo
  {pages} {7474} (\bibinfo {year} {2022})}\BibitemShut {NoStop}%
\bibitem [{\citenamefont {Lonardi}\ \emph {et~al.}(2023)\citenamefont
  {Lonardi}, \citenamefont {Facca}, \citenamefont {Putti},\ and\ \citenamefont
  {De~Bacco}}]{lonardi2021infrastructure}%
  \BibitemOpen
  \bibfield  {author} {\bibinfo {author} {\bibfnamefont {A.}~\bibnamefont
  {Lonardi}}, \bibinfo {author} {\bibfnamefont {E.}~\bibnamefont {Facca}},
  \bibinfo {author} {\bibfnamefont {M.}~\bibnamefont {Putti}},\ and\ \bibinfo
  {author} {\bibfnamefont {C.}~\bibnamefont {De~Bacco}},\ }\bibfield  {title}
  {\bibinfo {title} {Infrastructure adaptation and emergence of loops in
  network routing with time-dependent loads},\ }\href
  {https://doi.org/10.1103/PhysRevE.107.024302} {\bibfield  {journal} {\bibinfo
   {journal} {Phys. Rev. E}\ }\textbf {\bibinfo {volume} {107}},\ \bibinfo
  {pages} {024302} (\bibinfo {year} {2023})}\BibitemShut {NoStop}%
\bibitem [{\citenamefont {Demetci}\ \emph {et~al.}(2020)\citenamefont
  {Demetci}, \citenamefont {Santorella}, \citenamefont {Sandstede},
  \citenamefont {Noble},\ and\ \citenamefont {Singh}}]{demetci2020gromov}%
  \BibitemOpen
  \bibfield  {author} {\bibinfo {author} {\bibfnamefont {P.}~\bibnamefont
  {Demetci}}, \bibinfo {author} {\bibfnamefont {R.}~\bibnamefont {Santorella}},
  \bibinfo {author} {\bibfnamefont {B.}~\bibnamefont {Sandstede}}, \bibinfo
  {author} {\bibfnamefont {W.~S.}\ \bibnamefont {Noble}},\ and\ \bibinfo
  {author} {\bibfnamefont {R.}~\bibnamefont {Singh}},\ }\bibfield  {title}
  {\bibinfo {title} {Gromov-{W}asserstein optimal transport to align
  single-cell multi-omics data},\ }\bibfield  {journal} {\bibinfo  {journal}
  {bioRxiv}\ }\href {https://doi.org/10.1101/2020.04.28.066787}
  {10.1101/2020.04.28.066787} (\bibinfo {year} {2020})\BibitemShut {NoStop}%
\bibitem [{\citenamefont {Katifori}\ \emph {et~al.}(2010)\citenamefont
  {Katifori}, \citenamefont {Sz\"oll\ifmmode~\mbox{\H{o}}\else \H{o}\fi{}si},\
  and\ \citenamefont {Magnasco}}]{katifori2010damage}%
  \BibitemOpen
  \bibfield  {author} {\bibinfo {author} {\bibfnamefont {E.}~\bibnamefont
  {Katifori}}, \bibinfo {author} {\bibfnamefont {G.~J.}\ \bibnamefont
  {Sz\"oll\ifmmode~\mbox{\H{o}}\else \H{o}\fi{}si}},\ and\ \bibinfo {author}
  {\bibfnamefont {M.~O.}\ \bibnamefont {Magnasco}},\ }\bibfield  {title}
  {\bibinfo {title} {Damage and {F}luctuations {I}nduce {L}oops in {O}ptimal
  {T}ransport {N}etworks},\ }\href
  {https://doi.org/10.1103/PhysRevLett.104.048704} {\bibfield  {journal}
  {\bibinfo  {journal} {Phys. Rev. Lett.}\ }\textbf {\bibinfo {volume} {104}},\
  \bibinfo {pages} {048704} (\bibinfo {year} {2010})}\BibitemShut {NoStop}%
\bibitem [{\citenamefont {Werman}\ \emph {et~al.}(1985)\citenamefont {Werman},
  \citenamefont {Peleg},\ and\ \citenamefont {Rosenfeld}}]{werman1985distance}%
  \BibitemOpen
  \bibfield  {author} {\bibinfo {author} {\bibfnamefont {M.}~\bibnamefont
  {Werman}}, \bibinfo {author} {\bibfnamefont {S.}~\bibnamefont {Peleg}},\ and\
  \bibinfo {author} {\bibfnamefont {A.}~\bibnamefont {Rosenfeld}},\ }\bibfield
  {title} {\bibinfo {title} {A {D}istance {M}etric for {M}ultidimensional
  {H}istograms},\ }\href {https://doi.org/10.1016/0734-189X(85)90055-6}
  {\bibfield  {journal} {\bibinfo  {journal} {Computer {V}ision, {G}raphics,
  and {I}mage {P}rocessing}\ }\textbf {\bibinfo {volume} {32}},\ \bibinfo
  {pages} {328} (\bibinfo {year} {1985})}\BibitemShut {NoStop}%
\bibitem [{\citenamefont {Peleg}\ \emph {et~al.}(1989)\citenamefont {Peleg},
  \citenamefont {Werman},\ and\ \citenamefont {Rom}}]{peleg1989unified}%
  \BibitemOpen
  \bibfield  {author} {\bibinfo {author} {\bibfnamefont {S.}~\bibnamefont
  {Peleg}}, \bibinfo {author} {\bibfnamefont {M.}~\bibnamefont {Werman}},\ and\
  \bibinfo {author} {\bibfnamefont {H.}~\bibnamefont {Rom}},\ }\bibfield
  {title} {\bibinfo {title} {A unified approach to the change of resolution:
  space and gray-level},\ }\href {https://doi.org/10.1109/34.192468} {\bibfield
   {journal} {\bibinfo  {journal} {IEEE Transactions on Pattern Analysis and
  Machine Intelligence}\ }\textbf {\bibinfo {volume} {11}},\ \bibinfo {pages}
  {739} (\bibinfo {year} {1989})}\BibitemShut {NoStop}%
\bibitem [{\citenamefont {Rubner}\ \emph {et~al.}(1998)\citenamefont {Rubner},
  \citenamefont {Tomasi},\ and\ \citenamefont {Guibas}}]{rubner1998metric}%
  \BibitemOpen
  \bibfield  {author} {\bibinfo {author} {\bibfnamefont {Y.}~\bibnamefont
  {Rubner}}, \bibinfo {author} {\bibfnamefont {C.}~\bibnamefont {Tomasi}},\
  and\ \bibinfo {author} {\bibfnamefont {L.~J.}\ \bibnamefont {Guibas}},\
  }\bibfield  {title} {\bibinfo {title} {A metric for distributions with
  applications to image databases},\ }in\ \href
  {https://doi.org/10.1109/ICCV.1998.710701} {\emph {\bibinfo {booktitle}
  {Sixth International Conference on Computer Vision (IEEE Cat.
  No.98CH36271)}}}\ (\bibinfo {year} {1998})\ pp.\ \bibinfo {pages}
  {59--66}\BibitemShut {NoStop}%
\bibitem [{\citenamefont {Rubner}\ \emph {et~al.}(2000)\citenamefont {Rubner},
  \citenamefont {Tomasi},\ and\ \citenamefont {Guibas}}]{rubner2000earth}%
  \BibitemOpen
  \bibfield  {author} {\bibinfo {author} {\bibfnamefont {Y.}~\bibnamefont
  {Rubner}}, \bibinfo {author} {\bibfnamefont {C.}~\bibnamefont {Tomasi}},\
  and\ \bibinfo {author} {\bibfnamefont {L.~J.}\ \bibnamefont {Guibas}},\
  }\bibfield  {title} {\bibinfo {title} {{T}he {E}arth {M}over's {D}istance as
  a {M}etric for {I}mage {R}etrieval},\ }\href
  {https://doi.org/10.1023/A:1026543900054} {\bibfield  {journal} {\bibinfo
  {journal} {{I}nternational {J}ournal of {C}omputer {V}ision}\ }\textbf
  {\bibinfo {volume} {40}},\ \bibinfo {pages} {99} (\bibinfo {year}
  {2000})}\BibitemShut {NoStop}%
\bibitem [{\citenamefont {Baptista}\ and\ \citenamefont
  {De~Bacco}(2021)}]{baptista2020principled}%
  \BibitemOpen
  \bibfield  {author} {\bibinfo {author} {\bibfnamefont {D.}~\bibnamefont
  {Baptista}}\ and\ \bibinfo {author} {\bibfnamefont {C.}~\bibnamefont
  {De~Bacco}},\ }\bibfield  {title} {\bibinfo {title} {Principled network
  extraction from images},\ }\href {https://doi.org/10.1098/rsos.210025}
  {\bibfield  {journal} {\bibinfo  {journal} {R. Soc. open sci.}\ }\textbf
  {\bibinfo {volume} {8}},\ \bibinfo {pages} {210025} (\bibinfo {year}
  {2021})}\BibitemShut {NoStop}%
\bibitem [{\citenamefont {Peyré}\ and\ \citenamefont {Cuturi}(2019)}]{COTFNT}%
  \BibitemOpen
  \bibfield  {author} {\bibinfo {author} {\bibfnamefont {G.}~\bibnamefont
  {Peyré}}\ and\ \bibinfo {author} {\bibfnamefont {M.}~\bibnamefont
  {Cuturi}},\ }\bibfield  {title} {\bibinfo {title} {Computational {O}ptimal
  {T}ransport: With {A}pplications to {D}ata {S}cience},\ }\href
  {https://doi.org/10.1561/2200000073} {\bibfield  {journal} {\bibinfo
  {journal} {Foundations and {T}rends® in {M}achine {L}earning}\ }\textbf
  {\bibinfo {volume} {11}},\ \bibinfo {pages} {355} (\bibinfo {year}
  {2019})}\BibitemShut {NoStop}%
\bibitem [{\citenamefont {Koehl}\ \emph
  {et~al.}(2019{\natexlab{a}})\citenamefont {Koehl}, \citenamefont {Delarue},\
  and\ \citenamefont {Orland}}]{koehl_pre}%
  \BibitemOpen
  \bibfield  {author} {\bibinfo {author} {\bibfnamefont {P.}~\bibnamefont
  {Koehl}}, \bibinfo {author} {\bibfnamefont {M.}~\bibnamefont {Delarue}},\
  and\ \bibinfo {author} {\bibfnamefont {H.}~\bibnamefont {Orland}},\
  }\bibfield  {title} {\bibinfo {title} {Optimal transport at finite
  temperature},\ }\href {https://doi.org/10.1103/PhysRevE.100.013310}
  {\bibfield  {journal} {\bibinfo  {journal} {Phys. Rev. E}\ }\textbf {\bibinfo
  {volume} {100}},\ \bibinfo {pages} {013310} (\bibinfo {year}
  {2019}{\natexlab{a}})}\BibitemShut {NoStop}%
\bibitem [{\citenamefont {Aurell}\ \emph {et~al.}(2011)\citenamefont {Aurell},
  \citenamefont {Mej\'{\i}a-Monasterio},\ and\ \citenamefont
  {Muratore-Ginanneschi}}]{aurell2011optimal}%
  \BibitemOpen
  \bibfield  {author} {\bibinfo {author} {\bibfnamefont {E.}~\bibnamefont
  {Aurell}}, \bibinfo {author} {\bibfnamefont {C.}~\bibnamefont
  {Mej\'{\i}a-Monasterio}},\ and\ \bibinfo {author} {\bibfnamefont
  {P.}~\bibnamefont {Muratore-Ginanneschi}},\ }\bibfield  {title} {\bibinfo
  {title} {Optimal {P}rotocols and {O}ptimal {T}ransport in {S}tochastic
  {T}hermodynamics},\ }\href {https://doi.org/10.1103/PhysRevLett.106.250601}
  {\bibfield  {journal} {\bibinfo  {journal} {Phys. Rev. Lett.}\ }\textbf
  {\bibinfo {volume} {106}},\ \bibinfo {pages} {250601} (\bibinfo {year}
  {2011})}\BibitemShut {NoStop}%
\bibitem [{\citenamefont {Leite}\ and\ \citenamefont
  {De~Bacco}(2022)}]{leite2022revealing}%
  \BibitemOpen
  \bibfield  {author} {\bibinfo {author} {\bibfnamefont {D.}~\bibnamefont
  {Leite}}\ and\ \bibinfo {author} {\bibfnamefont {C.}~\bibnamefont
  {De~Bacco}},\ }\bibfield  {title} {\bibinfo {title} {Revealing the similarity
  between urban transportation networks and optimal transport-based
  infrastructures},\ }\href@noop {} {\bibfield  {journal} {\bibinfo  {journal}
  {arXiv preprint arXiv:2209.06751}\ } (\bibinfo {year} {2022})}\BibitemShut
  {NoStop}%
\bibitem [{\citenamefont {Baptista}\ \emph {et~al.}(2020)\citenamefont
  {Baptista}, \citenamefont {Leite}, \citenamefont {Facca}, \citenamefont
  {Putti},\ and\ \citenamefont {De~Bacco}}]{baptista2020network}%
  \BibitemOpen
  \bibfield  {author} {\bibinfo {author} {\bibfnamefont {D.}~\bibnamefont
  {Baptista}}, \bibinfo {author} {\bibfnamefont {D.}~\bibnamefont {Leite}},
  \bibinfo {author} {\bibfnamefont {E.}~\bibnamefont {Facca}}, \bibinfo
  {author} {\bibfnamefont {M.}~\bibnamefont {Putti}},\ and\ \bibinfo {author}
  {\bibfnamefont {C.}~\bibnamefont {De~Bacco}},\ }\bibfield  {title} {\bibinfo
  {title} {Network extraction by routing optimization},\ }\bibfield  {journal}
  {\bibinfo  {journal} {Scientific Reports}\ }\textbf {\bibinfo {volume}
  {10}},\ \href {https://doi.org/10.1038/s41598-020-77064-4}
  {10.1038/s41598-020-77064-4} (\bibinfo {year} {2020})\BibitemShut {NoStop}%
\bibitem [{\citenamefont {Ibrahim}\ \emph {et~al.}(2021)\citenamefont
  {Ibrahim}, \citenamefont {Lonardi},\ and\ \citenamefont
  {Bacco}}]{ibrahim2021optimal}%
  \BibitemOpen
  \bibfield  {author} {\bibinfo {author} {\bibfnamefont {A.~A.}\ \bibnamefont
  {Ibrahim}}, \bibinfo {author} {\bibfnamefont {A.}~\bibnamefont {Lonardi}},\
  and\ \bibinfo {author} {\bibfnamefont {C.~D.}\ \bibnamefont {Bacco}},\
  }\bibfield  {title} {\bibinfo {title} {Optimal {T}ransport in {M}ultilayer
  {N}etworks for {T}raffic {F}low {O}ptimization},\ }\bibfield  {journal}
  {\bibinfo  {journal} {Algorithms}\ }\textbf {\bibinfo {volume} {14}},\ \href
  {https://doi.org/10.3390/a14070189} {10.3390/a14070189} (\bibinfo {year}
  {2021})\BibitemShut {NoStop}%
\bibitem [{\citenamefont {Mondino}\ and\ \citenamefont
  {Suhr}(2022)}]{mondino2022optimal}%
  \BibitemOpen
  \bibfield  {author} {\bibinfo {author} {\bibfnamefont {A.}~\bibnamefont
  {Mondino}}\ and\ \bibinfo {author} {\bibfnamefont {S.}~\bibnamefont {Suhr}},\
  }\bibfield  {title} {\bibinfo {title} {An optimal transport formulation of
  the {E}instein equations of general relativity},\ }\bibfield  {journal}
  {\bibinfo  {journal} {Journal of the European Mathematical Society}\ }\href
  {https://doi.org/10.4171/JEMS/1188} {10.4171/JEMS/1188} (\bibinfo {year}
  {2022})\BibitemShut {NoStop}%
\bibitem [{\citenamefont {Grauman}\ and\ \citenamefont
  {Darrell}(2004)}]{grauman2004fast}%
  \BibitemOpen
  \bibfield  {author} {\bibinfo {author} {\bibfnamefont {K.}~\bibnamefont
  {Grauman}}\ and\ \bibinfo {author} {\bibfnamefont {T.}~\bibnamefont
  {Darrell}},\ }\bibfield  {title} {\bibinfo {title} {Fast contour matching
  using approximate earth mover's distance},\ }in\ \href
  {https://doi.org/10.1109/CVPR.2004.1315035} {\emph {\bibinfo {booktitle}
  {Proceedings of the 2004 IEEE Computer Society Conference on Computer Vision
  and Pattern Recognition, 2004. CVPR 2004}}},\ Vol.~\bibinfo {volume} {1}\
  (\bibinfo {year} {2004})\ pp.\ \bibinfo {pages} {I--I}\BibitemShut {NoStop}%
\bibitem [{\citenamefont {Cuturi}(2013)}]{cuturisinkhorn}%
  \BibitemOpen
  \bibfield  {author} {\bibinfo {author} {\bibfnamefont {M.}~\bibnamefont
  {Cuturi}},\ }\bibfield  {title} {\bibinfo {title} {Sinkhorn distances:
  Lightspeed computation of optimal transport},\ }in\ \href
  {https://proceedings.neurips.cc/paper/2013/file/af21d0c97db2e27e13572cbf59eb343d-Paper.pdf}
  {\emph {\bibinfo {booktitle} {Advances in Neural Information Processing
  Systems}}},\ Vol.~\bibinfo {volume} {26}\ (\bibinfo  {publisher} {Curran
  Associates, Inc.},\ \bibinfo {year} {2013})\ pp.\ \bibinfo {pages}
  {2292--2300}\BibitemShut {NoStop}%
\bibitem [{\citenamefont {Koehl}\ \emph
  {et~al.}(2019{\natexlab{b}})\citenamefont {Koehl}, \citenamefont {Delarue},\
  and\ \citenamefont {Orland}}]{koehl_prl}%
  \BibitemOpen
  \bibfield  {author} {\bibinfo {author} {\bibfnamefont {P.}~\bibnamefont
  {Koehl}}, \bibinfo {author} {\bibfnamefont {M.}~\bibnamefont {Delarue}},\
  and\ \bibinfo {author} {\bibfnamefont {H.}~\bibnamefont {Orland}},\
  }\bibfield  {title} {\bibinfo {title} {Statistical {P}hysics {A}pproach to
  the {O}ptimal {T}ransport {P}roblem},\ }\href
  {https://doi.org/10.1103/PhysRevLett.123.040603} {\bibfield  {journal}
  {\bibinfo  {journal} {Phys. Rev. Lett.}\ }\textbf {\bibinfo {volume} {123}},\
  \bibinfo {pages} {040603} (\bibinfo {year} {2019}{\natexlab{b}})}\BibitemShut
  {NoStop}%
\bibitem [{\citenamefont {Thorpe}\ \emph {et~al.}(2017)\citenamefont {Thorpe},
  \citenamefont {Park}, \citenamefont {Kolouri}, \citenamefont {Rohde},\ and\
  \citenamefont {Slep{\v{c}}ev}}]{thorpe2017transportation}%
  \BibitemOpen
  \bibfield  {author} {\bibinfo {author} {\bibfnamefont {M.}~\bibnamefont
  {Thorpe}}, \bibinfo {author} {\bibfnamefont {S.}~\bibnamefont {Park}},
  \bibinfo {author} {\bibfnamefont {S.}~\bibnamefont {Kolouri}}, \bibinfo
  {author} {\bibfnamefont {G.~K.}\ \bibnamefont {Rohde}},\ and\ \bibinfo
  {author} {\bibfnamefont {D.}~\bibnamefont {Slep{\v{c}}ev}},\ }\bibfield
  {title} {\bibinfo {title} {A transportation ${L}^{p}$ {D}istance for {S}ignal
  {A}nalysis},\ }\href {https://doi.org/10.1007/s10851-017-0726-4} {\bibfield
  {journal} {\bibinfo  {journal} {Journal of mathematical imaging and vision}\
  }\textbf {\bibinfo {volume} {59}},\ \bibinfo {pages} {187} (\bibinfo {year}
  {2017})}\BibitemShut {NoStop}%
\bibitem [{\citenamefont {Pele}\ and\ \citenamefont
  {Werman}(2008)}]{pele2008linear}%
  \BibitemOpen
  \bibfield  {author} {\bibinfo {author} {\bibfnamefont {O.}~\bibnamefont
  {Pele}}\ and\ \bibinfo {author} {\bibfnamefont {M.}~\bibnamefont {Werman}},\
  }\bibfield  {title} {\bibinfo {title} {{A} {L}inear {T}ime {H}istogram
  {M}etric for {I}mproved {SIFT} {M}atching},\ }in\ \href
  {https://doi.org/10.1007/978-3-540-88690-7_37} {\emph {\bibinfo {booktitle}
  {Computer Vision -- ECCV 2008}}}\ (\bibinfo  {publisher} {Springer Berlin
  Heidelberg},\ \bibinfo {address} {Berlin, Heidelberg},\ \bibinfo {year}
  {2008})\ pp.\ \bibinfo {pages} {495--508}\BibitemShut {NoStop}%
\bibitem [{\citenamefont {Pele}\ and\ \citenamefont
  {Werman}(2009)}]{pele2009fast}%
  \BibitemOpen
  \bibfield  {author} {\bibinfo {author} {\bibfnamefont {O.}~\bibnamefont
  {Pele}}\ and\ \bibinfo {author} {\bibfnamefont {M.}~\bibnamefont {Werman}},\
  }\bibfield  {title} {\bibinfo {title} {Fast and robust {E}arth {M}over's
  {D}istances},\ }in\ \href {https://doi.org/10.1109/ICCV.2009.5459199} {\emph
  {\bibinfo {booktitle} {2009 IEEE 12th International Conference on Computer
  Vision}}}\ (\bibinfo {year} {2009})\ pp.\ \bibinfo {pages}
  {460--467}\BibitemShut {NoStop}%
\bibitem [{\citenamefont {Villani}(2009)}]{villaniot}%
  \BibitemOpen
  \bibfield  {author} {\bibinfo {author} {\bibfnamefont {C.}~\bibnamefont
  {Villani}},\ }\href {https://doi.org/10.1007/978-3-540-71050-9} {\emph
  {\bibinfo {title} {Optimal transport: {O}ld and {N}ew}}},\ Vol.\ \bibinfo
  {volume} {338}\ (\bibinfo  {publisher} {Springer},\ \bibinfo {year}
  {2009})\BibitemShut {NoStop}%
\bibitem [{\citenamefont {Arjovsky}\ \emph {et~al.}(2017)\citenamefont
  {Arjovsky}, \citenamefont {Chintala},\ and\ \citenamefont
  {Bottou}}]{arjovsky2017wasserstein}%
  \BibitemOpen
  \bibfield  {author} {\bibinfo {author} {\bibfnamefont {M.}~\bibnamefont
  {Arjovsky}}, \bibinfo {author} {\bibfnamefont {S.}~\bibnamefont {Chintala}},\
  and\ \bibinfo {author} {\bibfnamefont {L.}~\bibnamefont {Bottou}},\
  }\bibfield  {title} {\bibinfo {title} {{W}asserstein {G}enerative
  {A}dversarial {N}etworks},\ }in\ \href
  {https://proceedings.mlr.press/v70/arjovsky17a.html} {\emph {\bibinfo
  {booktitle} {Proceedings of the 34th International Conference on Machine
  Learning}}},\ \bibinfo {series} {Proceedings of Machine Learning Research},
  Vol.~\bibinfo {volume} {70},\ \bibinfo {editor} {edited by\ \bibinfo {editor}
  {\bibfnamefont {D.}~\bibnamefont {Precup}}\ and\ \bibinfo {editor}
  {\bibfnamefont {Y.~W.}\ \bibnamefont {Teh}}}\ (\bibinfo  {publisher} {PMLR},\
  \bibinfo {year} {2017})\ pp.\ \bibinfo {pages} {214--223}\BibitemShut
  {NoStop}%
\bibitem [{\citenamefont {Lin}\ \emph {et~al.}(2019)\citenamefont {Lin},
  \citenamefont {Ho},\ and\ \citenamefont {Jordan}}]{lin2019efficient}%
  \BibitemOpen
  \bibfield  {author} {\bibinfo {author} {\bibfnamefont {T.}~\bibnamefont
  {Lin}}, \bibinfo {author} {\bibfnamefont {N.}~\bibnamefont {Ho}},\ and\
  \bibinfo {author} {\bibfnamefont {M.}~\bibnamefont {Jordan}},\ }\bibfield
  {title} {\bibinfo {title} {{O}n {E}fficient {O}ptimal {T}ransport: {A}n
  {A}nalysis of {G}reedy and {A}ccelerated {M}irror {D}escent {A}lgorithms},\
  }in\ \href {https://proceedings.mlr.press/v97/lin19a.html} {\emph {\bibinfo
  {booktitle} {Proceedings of the 36th International Conference on Machine
  Learning}}},\ \bibinfo {series} {Proceedings of Machine Learning Research},
  Vol.~\bibinfo {volume} {97}\ (\bibinfo  {publisher} {PMLR},\ \bibinfo {year}
  {2019})\ pp.\ \bibinfo {pages} {3982--3991}\BibitemShut {NoStop}%
\bibitem [{\citenamefont {Dvurechensky}\ \emph {et~al.}(2018)\citenamefont
  {Dvurechensky}, \citenamefont {Gasnikov},\ and\ \citenamefont
  {Kroshnin}}]{dvurechensky2018computational}%
  \BibitemOpen
  \bibfield  {author} {\bibinfo {author} {\bibfnamefont {P.}~\bibnamefont
  {Dvurechensky}}, \bibinfo {author} {\bibfnamefont {A.}~\bibnamefont
  {Gasnikov}},\ and\ \bibinfo {author} {\bibfnamefont {A.}~\bibnamefont
  {Kroshnin}},\ }\bibfield  {title} {\bibinfo {title} {{C}omputational
  {O}ptimal {T}ransport: {C}omplexity by {A}ccelerated {G}radient {D}escent
  {I}s {B}etter {T}han by {S}inkhorn’s {A}lgorithm},\ }in\ \href
  {https://proceedings.mlr.press/v80/dvurechensky18a.html} {\emph {\bibinfo
  {booktitle} {Proceedings of the 35th International Conference on Machine
  Learning}}},\ \bibinfo {series} {Proceedings of Machine Learning Research},
  Vol.~\bibinfo {volume} {80}\ (\bibinfo  {publisher} {PMLR},\ \bibinfo {year}
  {2018})\ pp.\ \bibinfo {pages} {1367--1376}\BibitemShut {NoStop}%
\bibitem [{\citenamefont {Banavar}\ \emph {et~al.}(2000)\citenamefont
  {Banavar}, \citenamefont {Colaiori}, \citenamefont {Flammini}, \citenamefont
  {Maritan},\ and\ \citenamefont {Rinaldo}}]{maritan}%
  \BibitemOpen
  \bibfield  {author} {\bibinfo {author} {\bibfnamefont {J.~R.}\ \bibnamefont
  {Banavar}}, \bibinfo {author} {\bibfnamefont {F.}~\bibnamefont {Colaiori}},
  \bibinfo {author} {\bibfnamefont {A.}~\bibnamefont {Flammini}}, \bibinfo
  {author} {\bibfnamefont {A.}~\bibnamefont {Maritan}},\ and\ \bibinfo {author}
  {\bibfnamefont {A.}~\bibnamefont {Rinaldo}},\ }\bibfield  {title} {\bibinfo
  {title} {Topology of the {F}ittest {T}ransportation {N}etwork},\ }\href
  {https://doi.org/10.1103/PhysRevLett.84.4745} {\bibfield  {journal} {\bibinfo
   {journal} {Phys. Rev. Lett.}\ }\textbf {\bibinfo {volume} {84}},\ \bibinfo
  {pages} {4745} (\bibinfo {year} {2000})}\BibitemShut {NoStop}%
\bibitem [{\citenamefont {Ronellenfitsch}\ and\ \citenamefont
  {Katifori}(2016)}]{ronellenfitsch2016global}%
  \BibitemOpen
  \bibfield  {author} {\bibinfo {author} {\bibfnamefont {H.}~\bibnamefont
  {Ronellenfitsch}}\ and\ \bibinfo {author} {\bibfnamefont {E.}~\bibnamefont
  {Katifori}},\ }\bibfield  {title} {\bibinfo {title} {Global {O}ptimization,
  {L}ocal {A}daptation, and the {R}ole of {G}rowth in {D}istribution
  {N}etworks},\ }\href {https://doi.org/10.1103/PhysRevLett.117.138301}
  {\bibfield  {journal} {\bibinfo  {journal} {Phys. Rev. Lett.}\ }\textbf
  {\bibinfo {volume} {117}},\ \bibinfo {pages} {138301} (\bibinfo {year}
  {2016})}\BibitemShut {NoStop}%
\bibitem [{\citenamefont {Hu}\ and\ \citenamefont
  {Cai}(2013)}]{hu2013adaptation}%
  \BibitemOpen
  \bibfield  {author} {\bibinfo {author} {\bibfnamefont {D.}~\bibnamefont
  {Hu}}\ and\ \bibinfo {author} {\bibfnamefont {D.}~\bibnamefont {Cai}},\
  }\bibfield  {title} {\bibinfo {title} {Adaptation and {O}ptimization of
  {B}iological {T}ransport {N}etworks},\ }\href
  {https://doi.org/10.1103/PhysRevLett.111.138701} {\bibfield  {journal}
  {\bibinfo  {journal} {Phys. Rev. Lett.}\ }\textbf {\bibinfo {volume} {111}},\
  \bibinfo {pages} {138701} (\bibinfo {year} {2013})}\BibitemShut {NoStop}%
\bibitem [{\citenamefont {Corson}(2010)}]{corson2010fluctuations}%
  \BibitemOpen
  \bibfield  {author} {\bibinfo {author} {\bibfnamefont {F.}~\bibnamefont
  {Corson}},\ }\bibfield  {title} {\bibinfo {title} {Fluctuations and
  {R}edundancy in {O}ptimal {T}ransport {N}etworks},\ }\href
  {https://doi.org/10.1103/PhysRevLett.104.048703} {\bibfield  {journal}
  {\bibinfo  {journal} {Phys. Rev. Lett.}\ }\textbf {\bibinfo {volume} {104}},\
  \bibinfo {pages} {048703} (\bibinfo {year} {2010})}\BibitemShut {NoStop}%
\bibitem [{\citenamefont {Lonardi}\ \emph {et~al.}(2021)\citenamefont
  {Lonardi}, \citenamefont {Facca}, \citenamefont {Putti},\ and\ \citenamefont
  {De~Bacco}}]{lonardi2021designing}%
  \BibitemOpen
  \bibfield  {author} {\bibinfo {author} {\bibfnamefont {A.}~\bibnamefont
  {Lonardi}}, \bibinfo {author} {\bibfnamefont {E.}~\bibnamefont {Facca}},
  \bibinfo {author} {\bibfnamefont {M.}~\bibnamefont {Putti}},\ and\ \bibinfo
  {author} {\bibfnamefont {C.}~\bibnamefont {De~Bacco}},\ }\bibfield  {title}
  {\bibinfo {title} {Designing optimal networks for multicommodity transport
  problem},\ }\href {https://doi.org/10.1103/PhysRevResearch.3.043010}
  {\bibfield  {journal} {\bibinfo  {journal} {Phys. Rev. Research}\ }\textbf
  {\bibinfo {volume} {3}},\ \bibinfo {pages} {043010} (\bibinfo {year}
  {2021})}\BibitemShut {NoStop}%
\bibitem [{\citenamefont {Bonifaci}\ \emph {et~al.}(2022)\citenamefont
  {Bonifaci}, \citenamefont {Facca}, \citenamefont {Folz}, \citenamefont
  {Karrenbauer}, \citenamefont {Kolev}, \citenamefont {Mehlhorn}, \citenamefont
  {Morigi}, \citenamefont {Shahkarami},\ and\ \citenamefont
  {Vermande}}]{bonifaci2022physarum}%
  \BibitemOpen
  \bibfield  {author} {\bibinfo {author} {\bibfnamefont {V.}~\bibnamefont
  {Bonifaci}}, \bibinfo {author} {\bibfnamefont {E.}~\bibnamefont {Facca}},
  \bibinfo {author} {\bibfnamefont {F.}~\bibnamefont {Folz}}, \bibinfo {author}
  {\bibfnamefont {A.}~\bibnamefont {Karrenbauer}}, \bibinfo {author}
  {\bibfnamefont {P.}~\bibnamefont {Kolev}}, \bibinfo {author} {\bibfnamefont
  {K.}~\bibnamefont {Mehlhorn}}, \bibinfo {author} {\bibfnamefont
  {G.}~\bibnamefont {Morigi}}, \bibinfo {author} {\bibfnamefont
  {G.}~\bibnamefont {Shahkarami}},\ and\ \bibinfo {author} {\bibfnamefont
  {Q.}~\bibnamefont {Vermande}},\ }\bibfield  {title} {\bibinfo {title}
  {Physarum-inspired multi-commodity flow dynamics},\ }\bibfield  {journal}
  {\bibinfo  {journal} {Theoretical Computer Science}\ }\href
  {https://doi.org/10.1016/j.tcs.2022.02.001} {10.1016/j.tcs.2022.02.001}
  (\bibinfo {year} {2022})\BibitemShut {NoStop}%
\bibitem [{\citenamefont {Bonifaci}\ \emph {et~al.}(2012)\citenamefont
  {Bonifaci}, \citenamefont {Mehlhorn},\ and\ \citenamefont
  {Varma}}]{bonifaci2012physarum}%
  \BibitemOpen
  \bibfield  {author} {\bibinfo {author} {\bibfnamefont {V.}~\bibnamefont
  {Bonifaci}}, \bibinfo {author} {\bibfnamefont {K.}~\bibnamefont {Mehlhorn}},\
  and\ \bibinfo {author} {\bibfnamefont {G.}~\bibnamefont {Varma}},\ }\bibfield
   {title} {\bibinfo {title} {Physarum can compute shortest paths},\ }\href
  {https://doi.org/10.1016/j.jtbi.2012.06.017} {\bibfield  {journal} {\bibinfo
  {journal} {Journal of Theoretical Biology}\ }\textbf {\bibinfo {volume}
  {309}},\ \bibinfo {pages} {121 } (\bibinfo {year} {2012})}\BibitemShut
  {NoStop}%
\bibitem [{\citenamefont {Rinaldo}\ \emph {et~al.}(1993)\citenamefont
  {Rinaldo}, \citenamefont {Rodriguez-Iturbe}, \citenamefont {Rigon},
  \citenamefont {Ijjasz-Vasquez},\ and\ \citenamefont
  {Bras}}]{rinaldo1993self}%
  \BibitemOpen
  \bibfield  {author} {\bibinfo {author} {\bibfnamefont {A.}~\bibnamefont
  {Rinaldo}}, \bibinfo {author} {\bibfnamefont {I.}~\bibnamefont
  {Rodriguez-Iturbe}}, \bibinfo {author} {\bibfnamefont {R.}~\bibnamefont
  {Rigon}}, \bibinfo {author} {\bibfnamefont {E.}~\bibnamefont
  {Ijjasz-Vasquez}},\ and\ \bibinfo {author} {\bibfnamefont {R.~L.}\
  \bibnamefont {Bras}},\ }\bibfield  {title} {\bibinfo {title} {Self-organized
  fractal river networks},\ }\href {https://doi.org/10.1103/PhysRevLett.70.822}
  {\bibfield  {journal} {\bibinfo  {journal} {Phys. Rev. Lett.}\ }\textbf
  {\bibinfo {volume} {70}},\ \bibinfo {pages} {822} (\bibinfo {year}
  {1993})}\BibitemShut {NoStop}%
\bibitem [{\citenamefont {Barab\'asi}(1996)}]{barabasi1996invasion}%
  \BibitemOpen
  \bibfield  {author} {\bibinfo {author} {\bibfnamefont {A.-L.}\ \bibnamefont
  {Barab\'asi}},\ }\bibfield  {title} {\bibinfo {title} {Invasion {P}ercolation
  and {G}lobal {O}ptimization},\ }\href
  {https://doi.org/10.1103/PhysRevLett.76.3750} {\bibfield  {journal} {\bibinfo
   {journal} {Phys. Rev. Lett.}\ }\textbf {\bibinfo {volume} {76}},\ \bibinfo
  {pages} {3750} (\bibinfo {year} {1996})}\BibitemShut {NoStop}%
\bibitem [{\citenamefont {Santambrogio}(2007)}]{santambrogio2007optimal}%
  \BibitemOpen
  \bibfield  {author} {\bibinfo {author} {\bibfnamefont {F.}~\bibnamefont
  {Santambrogio}},\ }\bibfield  {title} {\bibinfo {title} {Optimal channel
  networks, landscape function and branched transport},\ }\href
  {https://doi.org/10.4171/IFB/160} {\bibfield  {journal} {\bibinfo  {journal}
  {Interfaces and {F}ree {B}oundaries}\ }\textbf {\bibinfo {volume} {9}},\
  \bibinfo {pages} {149} (\bibinfo {year} {2007})}\BibitemShut {NoStop}%
\bibitem [{\citenamefont {Ibrahim}\ \emph {et~al.}(2022)\citenamefont
  {Ibrahim}, \citenamefont {Leite},\ and\ \citenamefont {{De
  Bacco}}}]{ibrahim2022sustainable}%
  \BibitemOpen
  \bibfield  {author} {\bibinfo {author} {\bibfnamefont {A.~A.}\ \bibnamefont
  {Ibrahim}}, \bibinfo {author} {\bibfnamefont {D.}~\bibnamefont {Leite}},\
  and\ \bibinfo {author} {\bibfnamefont {C.}~\bibnamefont {{De Bacco}}},\
  }\bibfield  {title} {\bibinfo {title} {Sustainable optimal transport in
  multilayer networks},\ }\href
  {https://doi.org/https://doi.org/10.1103/PhysRevE.105.064302} {\bibfield
  {journal} {\bibinfo  {journal} {Physical Review E}\ }\textbf {\bibinfo
  {volume} {105}},\ \bibinfo {pages} {064302} (\bibinfo {year}
  {2022})}\BibitemShut {NoStop}%
\bibitem [{\citenamefont {Facca}\ and\ \citenamefont
  {Benzi}(2021)}]{facca2020fast}%
  \BibitemOpen
  \bibfield  {author} {\bibinfo {author} {\bibfnamefont {E.}~\bibnamefont
  {Facca}}\ and\ \bibinfo {author} {\bibfnamefont {M.}~\bibnamefont {Benzi}},\
  }\bibfield  {title} {\bibinfo {title} {Fast {I}terative {S}olution of the
  {O}ptimal {T}ransport {P}roblem on {G}raphs},\ }\href
  {https://doi.org/10.1137/20M137015X} {\bibfield  {journal} {\bibinfo
  {journal} {SIAM Journal on Scientific Computing}\ }\textbf {\bibinfo {volume}
  {43}},\ \bibinfo {pages} {A2295} (\bibinfo {year} {2021})}\BibitemShut
  {NoStop}%
\bibitem [{\citenamefont {Flamary}\ \emph {et~al.}(2021)\citenamefont
  {Flamary}, \citenamefont {Courty}, \citenamefont {Gramfort}, \citenamefont
  {Alaya}, \citenamefont {Boisbunon}, \citenamefont {Chambon}, \citenamefont
  {Chapel}, \citenamefont {Corenflos}, \citenamefont {Fatras}, \citenamefont
  {Fournier}, \citenamefont {Gautheron}, \citenamefont {Gayraud}, \citenamefont
  {Janati}, \citenamefont {Rakotomamonjy}, \citenamefont {Redko}, \citenamefont
  {Rolet}, \citenamefont {Schutz}, \citenamefont {Seguy}, \citenamefont
  {Sutherland}, \citenamefont {Tavenard}, \citenamefont {Tong},\ and\
  \citenamefont {Vayer}}]{flamary2021pot}%
  \BibitemOpen
  \bibfield  {author} {\bibinfo {author} {\bibfnamefont {R.}~\bibnamefont
  {Flamary}}, \bibinfo {author} {\bibfnamefont {N.}~\bibnamefont {Courty}},
  \bibinfo {author} {\bibfnamefont {A.}~\bibnamefont {Gramfort}}, \bibinfo
  {author} {\bibfnamefont {M.~Z.}\ \bibnamefont {Alaya}}, \bibinfo {author}
  {\bibfnamefont {A.}~\bibnamefont {Boisbunon}}, \bibinfo {author}
  {\bibfnamefont {S.}~\bibnamefont {Chambon}}, \bibinfo {author} {\bibfnamefont
  {L.}~\bibnamefont {Chapel}}, \bibinfo {author} {\bibfnamefont
  {A.}~\bibnamefont {Corenflos}}, \bibinfo {author} {\bibfnamefont
  {K.}~\bibnamefont {Fatras}}, \bibinfo {author} {\bibfnamefont
  {N.}~\bibnamefont {Fournier}}, \bibinfo {author} {\bibfnamefont
  {L.}~\bibnamefont {Gautheron}}, \bibinfo {author} {\bibfnamefont {N.~T.}\
  \bibnamefont {Gayraud}}, \bibinfo {author} {\bibfnamefont {H.}~\bibnamefont
  {Janati}}, \bibinfo {author} {\bibfnamefont {A.}~\bibnamefont
  {Rakotomamonjy}}, \bibinfo {author} {\bibfnamefont {I.}~\bibnamefont
  {Redko}}, \bibinfo {author} {\bibfnamefont {A.}~\bibnamefont {Rolet}},
  \bibinfo {author} {\bibfnamefont {A.}~\bibnamefont {Schutz}}, \bibinfo
  {author} {\bibfnamefont {V.}~\bibnamefont {Seguy}}, \bibinfo {author}
  {\bibfnamefont {D.~J.}\ \bibnamefont {Sutherland}}, \bibinfo {author}
  {\bibfnamefont {R.}~\bibnamefont {Tavenard}}, \bibinfo {author}
  {\bibfnamefont {A.}~\bibnamefont {Tong}},\ and\ \bibinfo {author}
  {\bibfnamefont {T.}~\bibnamefont {Vayer}},\ }\bibfield  {title} {\bibinfo
  {title} {{POT}: {P}ython {O}ptimal {T}ransport},\ }\href
  {http://jmlr.org/papers/v22/20-451.html} {\bibfield  {journal} {\bibinfo
  {journal} {Journal of Machine Learning Research}\ }\textbf {\bibinfo {volume}
  {22}},\ \bibinfo {pages} {1} (\bibinfo {year} {2021})}\BibitemShut {NoStop}%
\bibitem [{\citenamefont {Schmitzer}(2019)}]{schmitzer2019stabilized}%
  \BibitemOpen
  \bibfield  {author} {\bibinfo {author} {\bibfnamefont {B.}~\bibnamefont
  {Schmitzer}},\ }\bibfield  {title} {\bibinfo {title} {Stabilized {S}parse
  {S}caling {A}lgorithms for {E}ntropy {R}egularized {T}ransport {P}roblems},\
  }\href {https://doi.org/10.1137/16M1106018} {\bibfield  {journal} {\bibinfo
  {journal} {SIAM Journal on Scientific Computing}\ }\textbf {\bibinfo {volume}
  {41}},\ \bibinfo {pages} {A1443} (\bibinfo {year} {2019})}\BibitemShut
  {NoStop}%
\bibitem [{\citenamefont {Seeland}\ \emph {et~al.}(2017)\citenamefont
  {Seeland}, \citenamefont {Rzanny}, \citenamefont {Alaqraa}, \citenamefont
  {Wäldchen},\ and\ \citenamefont {Mäder}}]{seeland2017plant}%
  \BibitemOpen
  \bibfield  {author} {\bibinfo {author} {\bibfnamefont {M.}~\bibnamefont
  {Seeland}}, \bibinfo {author} {\bibfnamefont {M.}~\bibnamefont {Rzanny}},
  \bibinfo {author} {\bibfnamefont {N.}~\bibnamefont {Alaqraa}}, \bibinfo
  {author} {\bibfnamefont {J.}~\bibnamefont {Wäldchen}},\ and\ \bibinfo
  {author} {\bibfnamefont {P.}~\bibnamefont {Mäder}},\ }\href
  {https://doi.org/10.7910/DVN/QDHYST} {\bibinfo {title} {{Jena Flowers 30
  Dataset}}} (\bibinfo {year} {2017})\BibitemShut {NoStop}%
\bibitem [{\citenamefont {Macanh{\~a}}\ \emph {et~al.}(2018)\citenamefont
  {Macanh{\~a}}, \citenamefont {Eler}, \citenamefont {Garcia},\ and\
  \citenamefont {Junior}}]{alves2018handwritten}%
  \BibitemOpen
  \bibfield  {author} {\bibinfo {author} {\bibfnamefont {P.~A.}\ \bibnamefont
  {Macanh{\~a}}}, \bibinfo {author} {\bibfnamefont {D.~M.}\ \bibnamefont
  {Eler}}, \bibinfo {author} {\bibfnamefont {R.~E.}\ \bibnamefont {Garcia}},\
  and\ \bibinfo {author} {\bibfnamefont {W.~E.~M.}\ \bibnamefont {Junior}},\
  }\bibfield  {title} {\bibinfo {title} {Handwritten feature descriptor methods
  applied to fruit classification},\ }in\ \href
  {https://doi.org/10.1007/978-3-319-54978-1_87} {\emph {\bibinfo {booktitle}
  {Information Technology - New Generations}}}\ (\bibinfo  {publisher}
  {Springer International Publishing},\ \bibinfo {address} {Cham},\ \bibinfo
  {year} {2018})\ pp.\ \bibinfo {pages} {699--705}\BibitemShut {NoStop}%
\bibitem [{\citenamefont {Trottenberg}\ \emph {et~al.}(2000)\citenamefont
  {Trottenberg}, \citenamefont {Oosterlee},\ and\ \citenamefont
  {Schuller}}]{trottenberg2000multigrid}%
  \BibitemOpen
  \bibfield  {author} {\bibinfo {author} {\bibfnamefont {U.}~\bibnamefont
  {Trottenberg}}, \bibinfo {author} {\bibfnamefont {C.~W.}\ \bibnamefont
  {Oosterlee}},\ and\ \bibinfo {author} {\bibfnamefont {A.}~\bibnamefont
  {Schuller}},\ }\href@noop {} {\emph {\bibinfo {title} {Multigrid}}}\
  (\bibinfo  {publisher} {Elsevier},\ \bibinfo {year} {2000})\BibitemShut
  {NoStop}%
\bibitem [{\citenamefont {Eisenberger}\ \emph {et~al.}(2022)\citenamefont
  {Eisenberger}, \citenamefont {Toker}, \citenamefont {Leal-Taix\'e},
  \citenamefont {Bernard},\ and\ \citenamefont
  {Cremers}}]{eisenberger2022unified}%
  \BibitemOpen
  \bibfield  {author} {\bibinfo {author} {\bibfnamefont {M.}~\bibnamefont
  {Eisenberger}}, \bibinfo {author} {\bibfnamefont {A.}~\bibnamefont {Toker}},
  \bibinfo {author} {\bibfnamefont {L.}~\bibnamefont {Leal-Taix\'e}}, \bibinfo
  {author} {\bibfnamefont {F.}~\bibnamefont {Bernard}},\ and\ \bibinfo {author}
  {\bibfnamefont {D.}~\bibnamefont {Cremers}},\ }\bibfield  {title} {\bibinfo
  {title} {A unified framework for implicit sinkhorn differentiation},\ }in\
  \href@noop {} {\emph {\bibinfo {booktitle} {Proceedings of the IEEE/CVF
  Conference on Computer Vision and Pattern Recognition (CVPR)}}}\ (\bibinfo
  {year} {2022})\ pp.\ \bibinfo {pages} {509--518}\BibitemShut {NoStop}%
\bibitem [{git(2022)}]{git_repo}%
  \BibitemOpen
  \href@noop {} {\bibinfo {title} {{MODI} ({O}pen {S}ource code
  implementation)}} (\bibinfo {year} {2022}),\ \bibinfo {note}
  {\href{https://github.com/aleable/MODI}{https://github.com/aleable/MODI}}\BibitemShut
  {NoStop}%
\bibitem [{\citenamefont {Shepard}(1987)}]{shepard1987Toward}%
  \BibitemOpen
  \bibfield  {author} {\bibinfo {author} {\bibfnamefont {R.~N.}\ \bibnamefont
  {Shepard}},\ }\bibfield  {title} {\bibinfo {title} {Toward a {U}niversal
  {L}aw of {G}eneralization for {P}sychological {S}cience},\ }\href
  {https://doi.org/10.1126/science.3629243} {\bibfield  {journal} {\bibinfo
  {journal} {Science}\ }\textbf {\bibinfo {volume} {237}},\ \bibinfo {pages}
  {1317} (\bibinfo {year} {1987})}\BibitemShut {NoStop}%
\bibitem [{\citenamefont {Poynton}(1997)}]{skimage}%
  \BibitemOpen
  \bibfield  {author} {\bibinfo {author} {\bibfnamefont {C.}~\bibnamefont
  {Poynton}},\ }\bibfield  {title} {\bibinfo {title} {{F}requently {A}sked
  {Q}uestions about {C}olor.},\ }\href@noop {} {\  (\bibinfo {year}
  {1997})}\BibitemShut {NoStop}%
\bibitem [{\citenamefont {Sinkhorn}(1964)}]{sinkhorn1964relationship}%
  \BibitemOpen
  \bibfield  {author} {\bibinfo {author} {\bibfnamefont {R.}~\bibnamefont
  {Sinkhorn}},\ }\bibfield  {title} {\bibinfo {title} {A relationship between
  arbitrary positive matrices and doubly stochastic matrices},\ }\href
  {https://doi.org/10.1214/aoms/1177703591} {\bibfield  {journal} {\bibinfo
  {journal} {The annals of mathematical statistics}\ }\textbf {\bibinfo
  {volume} {35}},\ \bibinfo {pages} {876} (\bibinfo {year} {1964})}\BibitemShut
  {NoStop}%
\bibitem [{\citenamefont {Knopp}\ and\ \citenamefont
  {Sinkhorn}(1967)}]{knopp_sinkhorn}%
  \BibitemOpen
  \bibfield  {author} {\bibinfo {author} {\bibfnamefont {P.}~\bibnamefont
  {Knopp}}\ and\ \bibinfo {author} {\bibfnamefont {R.}~\bibnamefont
  {Sinkhorn}},\ }\bibfield  {title} {\bibinfo {title} {Concerning nonnegative
  matrices and doubly stochastic matrices},\ }\href
  {https://doi.org/pjm/1102992505} {\bibfield  {journal} {\bibinfo  {journal}
  {Pacific Journal of Mathematics}\ }\textbf {\bibinfo {volume} {21}},\
  \bibinfo {pages} {343 } (\bibinfo {year} {1967})}\BibitemShut {NoStop}%
\bibitem [{\citenamefont {Genevay}\ \emph {et~al.}(2016)\citenamefont
  {Genevay}, \citenamefont {Cuturi}, \citenamefont {Peyr\'{e}},\ and\
  \citenamefont {Bach}}]{genevay2016stochastic}%
  \BibitemOpen
  \bibfield  {author} {\bibinfo {author} {\bibfnamefont {A.}~\bibnamefont
  {Genevay}}, \bibinfo {author} {\bibfnamefont {M.}~\bibnamefont {Cuturi}},
  \bibinfo {author} {\bibfnamefont {G.}~\bibnamefont {Peyr\'{e}}},\ and\
  \bibinfo {author} {\bibfnamefont {F.}~\bibnamefont {Bach}},\ }\bibfield
  {title} {\bibinfo {title} {Stochastic optimization for large-scale optimal
  transport},\ }in\ \href
  {https://proceedings.neurips.cc/paper/2016/file/2a27b8144ac02f67687f76782a3b5d8f-Paper.pdf}
  {\emph {\bibinfo {booktitle} {Advances in Neural Information Processing
  Systems}}},\ Vol.~\bibinfo {volume} {29},\ \bibinfo {editor} {edited by\
  \bibinfo {editor} {\bibfnamefont {D.}~\bibnamefont {Lee}}, \bibinfo {editor}
  {\bibfnamefont {M.}~\bibnamefont {Sugiyama}}, \bibinfo {editor}
  {\bibfnamefont {U.}~\bibnamefont {Luxburg}}, \bibinfo {editor} {\bibfnamefont
  {I.}~\bibnamefont {Guyon}},\ and\ \bibinfo {editor} {\bibfnamefont
  {R.}~\bibnamefont {Garnett}}}\ (\bibinfo  {publisher} {Curran Associates,
  Inc.},\ \bibinfo {year} {2016})\BibitemShut {NoStop}%
\bibitem [{\citenamefont {Altschuler}\ \emph {et~al.}(2017)\citenamefont
  {Altschuler}, \citenamefont {Niles-Weed},\ and\ \citenamefont
  {Rigollet}}]{altschuler2017near}%
  \BibitemOpen
  \bibfield  {author} {\bibinfo {author} {\bibfnamefont {J.}~\bibnamefont
  {Altschuler}}, \bibinfo {author} {\bibfnamefont {J.}~\bibnamefont
  {Niles-Weed}},\ and\ \bibinfo {author} {\bibfnamefont {P.}~\bibnamefont
  {Rigollet}},\ }\bibfield  {title} {\bibinfo {title} {Near-linear time
  approximation algorithms for optimal transport via sinkhorn iteration},\ }in\
  \href
  {https://proceedings.neurips.cc/paper/2017/file/491442df5f88c6aa018e86dac21d3606-Paper.pdf}
  {\emph {\bibinfo {booktitle} {Advances in Neural Information Processing
  Systems}}},\ Vol.~\bibinfo {volume} {30},\ \bibinfo {editor} {edited by\
  \bibinfo {editor} {\bibfnamefont {I.}~\bibnamefont {Guyon}}, \bibinfo
  {editor} {\bibfnamefont {U.~V.}\ \bibnamefont {Luxburg}}, \bibinfo {editor}
  {\bibfnamefont {S.}~\bibnamefont {Bengio}}, \bibinfo {editor} {\bibfnamefont
  {H.}~\bibnamefont {Wallach}}, \bibinfo {editor} {\bibfnamefont
  {R.}~\bibnamefont {Fergus}}, \bibinfo {editor} {\bibfnamefont
  {S.}~\bibnamefont {Vishwanathan}},\ and\ \bibinfo {editor} {\bibfnamefont
  {R.}~\bibnamefont {Garnett}}}\ (\bibinfo  {publisher} {Curran Associates,
  Inc.},\ \bibinfo {year} {2017})\BibitemShut {NoStop}%
\bibitem [{\citenamefont {Chizat}\ \emph {et~al.}(2018)\citenamefont {Chizat},
  \citenamefont {Peyr{\'e}}, \citenamefont {Schmitzer},\ and\ \citenamefont
  {Vialard}}]{chizat2018scaling}%
  \BibitemOpen
  \bibfield  {author} {\bibinfo {author} {\bibfnamefont {L.}~\bibnamefont
  {Chizat}}, \bibinfo {author} {\bibfnamefont {G.}~\bibnamefont {Peyr{\'e}}},
  \bibinfo {author} {\bibfnamefont {B.}~\bibnamefont {Schmitzer}},\ and\
  \bibinfo {author} {\bibfnamefont {F.-X.}\ \bibnamefont {Vialard}},\
  }\bibfield  {title} {\bibinfo {title} {Scaling algorithms for unbalanced
  optimal transport problems},\ }\href {https://doi.org/10.1090/mcom/3303}
  {\bibfield  {journal} {\bibinfo  {journal} {Mathematics of Computation}\
  }\textbf {\bibinfo {volume} {87}},\ \bibinfo {pages} {2563} (\bibinfo {year}
  {2018})}\BibitemShut {NoStop}%
\end{thebibliography}%

%


\onecolumngrid
\mbox{}
\clearpage
\newpage

\setcounter{equation}{0}
\setcounter{figure}{0}
\setcounter{section}{0}
\setcounter{table}{0}
\setcounter{page}{1}
\makeatletter
\renewcommand{\theequation}{S\arabic{equation}}
\renewcommand{\thefigure}{S\arabic{figure}}
\renewcommand{\thetable}{S\arabic{table}}

\title{Immiscible Color Flows in Optimal Transport Networks for Image Classification: Supplemental Material (SM)}

\maketitle
\onecolumngrid\vspace{-1cm}

\section{Construction of the network}

\begin{figure}[b]
    \centering
    \includegraphics[width=\linewidth]{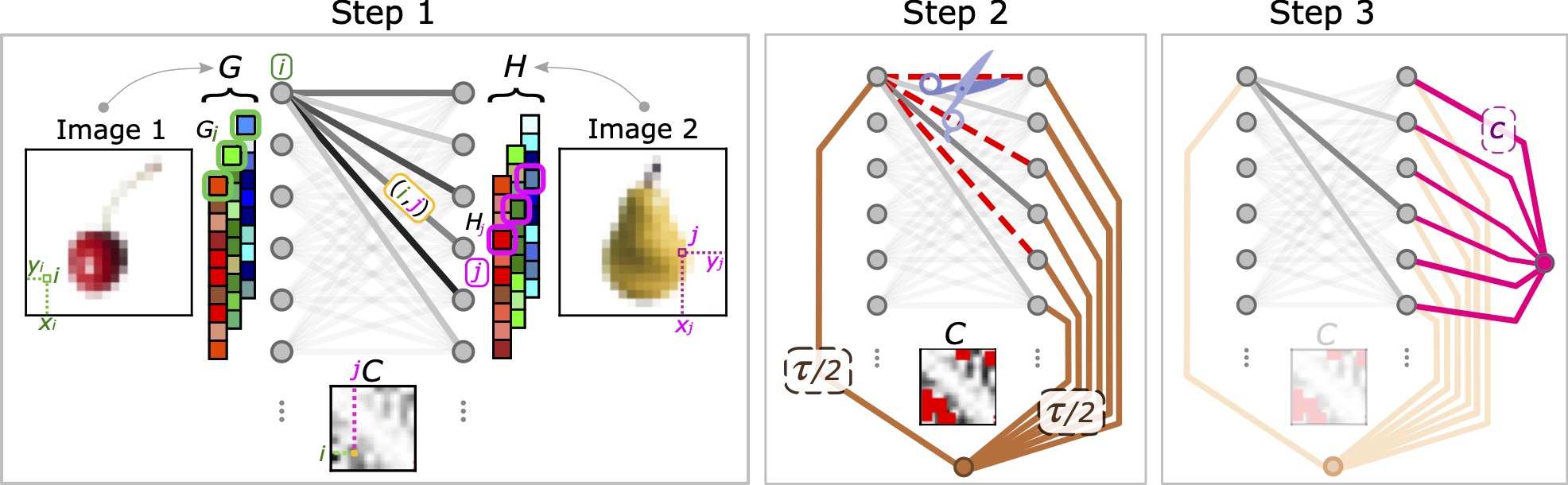}
    \caption{{Detailed construction of transport networks. Step 1: conversion of colored images to tensors and construction of the first complete bipartite graph. Step 2: trimming of expensive edges and addition of the transshipment node, $u_1$, with its links (in brown). Step 3: relaxation of mass balance with the addition of the second auxiliary node, $u_2$, together with its links (in magenta). }}
    \label{apxfig:model_construction_panel}
\end{figure}

Here, we extensively explain the procedure used for constructing the transport network $K$. All the essential steps are schematically drawn in \Cref{apxfig:model_construction_panel}, and are as follows.

\begin{itemize} 
    \item[Step 1.] Initially, a pair of images (Image 1, Image 2), is given as a couple of multidimensional arrays of dimensions $(w_i,h_i,M=3)$, with $i = 1,2$. We denote with $w$ images' widths and with $h$ their heights. The third dimension has size $M=3$, and corresponds to the three RGB color channels. The color channels are flattened to obtain the tensors $G$ and $H$,  the first for Image 1, and the second for Image 2. In detail, each channel is vectorized to have dimension $m \times 1$, with $m = w_1 \cdot h_1$,  for $g^a$ (resp. $n \times 1$, with $n = w_2 \cdot h_2$, for $h^a$), which are inflows and outflows of our multicommodity dynamics. In this way, the tensors $G$ and $H$, which are obtained stacking $g^a$ and $h^a$, have size $m \times M$ and $n \times M$. Entries of $G$ and $H$ are in standard RGB encoding, hence they are integers ranging from $0$ to $255$.
    
    To obtain the transport network $K$, we first generate a complete bipartite graph between $m+n = |V_1| + |V_2|$ nodes, the first $m = |V_1|$ are the pixels of Image 1, and the other $n = |V_2|$ are those of Image 2. We then assign a cost to each edge of this graph using both information given by pixels' locations, and by images' colors. In particular, we define:
    \begin{align}
        \label{apxeqn:C}
        C_{e}(\theta) &= (1-\theta)Y_e + \theta X_e \quad \forall e = (i,j)\\
        \label{apxeqn:V}
        Y_{e=(i,j)} &= || v_i - v_j ||_2 = \sqrt{(x_i - x_j)^2 + (y_i - y_j)^2}\\
        \label{apxeqn:X}
        X_{e=(i,j)} &= || G_i - H_j ||_1 = \sum_{a=1}^{M=3} \left| G_{ia} - H_{ja} \right|
    \end{align}
    for each $i$ pixel of Image 1, and $j$ pixel of Image 2. Terms $Y_e$ in \cref{apxeqn:V} contain the Euclidean distance between any pair of pixels, whose horizontal and vertical coordinates are stored in vectors $v = (x, y)$. Instead, $X_e$ contributes with colors to edges' costs. We model the effect of colors taking, in  \cref{apxeqn:X}, the 1-norm between arrays $G_i,H_j$, containing the RGB intensities in $i$ (pixel of Image 1) and $j$ (pixel of Image 2). Both $X_e$ and $Y_e$ have been opportunely rescaled in the range $[0,1]$. Lastly, we use the scalar parameter $0 \leq \theta \leq 1$ to weigh $Y_e$ and $X_e$ in a convex combination, in \cref{apxeqn:C}.

    \item[Step 2.] Once Step 1 is complete, and a cost $C_e$ is assigned to each edge of the complete bipartite graph between the two images, we implement a trimming procedure similar to that of \cite{pele2008linear,pele2009fast} to cut highly expensive links. In particular, we trim all edges $e$ that have cost $C_e > \tau$, where $\tau > 0$ is a threshold fixed a priori.  The links between $V_1$ and $V_2$ that do not get cut make up the set $E_{12}$. We then add a first transshipment node, $u_1$, to the network, and connect it with $m+n$ links to the sets $V_1$ and $V_2$. Each transshipment link is assigned a fixed cost $C_e = \tau/2$. This implies that one needs to pay a total cost of $\tau$  to transport mass from a node of Image 1 (in $V_1$) to one of Image 2 (in $V_2$),  when traversing transshipment links.
    
    There are several benefits in thresholding for the cost: (i) from a purely intuitive standpoint, humans perceive distances as saturated distances \cite{shepard1987Toward}; (ii) many natural color distributions are noisy and heavy-tailed, thus thresholding permits to assign a fixed cost to outliars; (iii) thresholded distances induce a $W_1$ distance between distributions in standard unicommodity OT problems \cite{pele2009fast}. More practically, thresholding improves accuracy and speed of OT \cite{pele2009fast} (see also the Computational Cost Section in this SM).
    
    \item[Step 3.] The last step required to obtain $K$ is the introduction of a second auxiliary node, $u_2$, together with its edges, to relax mass balance. In detail, in a standard OT setting $\sum_i G_{ia} = \sum_j H_{ja} = \Lambda^a > 0$ holds $\forall a = 1,\dots, M$, i.e., two histograms to be transported belong to the same simplex of mass $\Lambda^a > 0$. We relax this constraint permitting $\sum_i G_{ia} \neq \sum_j H_{ja}$ and penalizing Eq. (1) (main text). Particularly,  we use a similar relaxation of that in \cite{pele2009fast}, which we generalize to the multicommodity setup:
\begin{multline}
    \label{apxeqn:relax}
    J^\star_\Gamma (G,H) = \min_{P \in \Pi(G,H)} J_\Gamma(G,H) \quad \overset{\text{Relaxation}}{\longrightarrow} \quad \min_{P} \left\{ J_\Gamma(G,H) + \alpha \textstyle\sum_a \left| \textstyle\sum_j H_{ja} - \textstyle\sum_i G_{ia} \right| \max_{e \in E_{12}} {C_e} \right\}.
\end{multline}
The intuition of \cref{apxeqn:relax} is that the OT problem is penalized proportionally to the net difference between the inflowing and the outflowing mass. Hence, two images whose colors strongly differ return a higher cost and, in a supervised classification task, are less likely to be assigned the same label. We fix $\alpha = 1/2$ as in \cite{pele2008linear}.

This penalization can be translated to the transport network with the addition of $n$ links, costing $c = \alpha \max_{e \in E_{12}} {C_e} = \max_{e \in E_{12}} {C_e}/2$, connected to $u_2$. The excess of mass $m^a= \sum_j H_{ja} - \sum_i G_{ia}$ given by each commodity is injected $u_2$ to guarantee that the whole system is isolated. With this expedient one recovers exactly the relaxed OT formulation in \cref{apxeqn:relax}. In fact, all the transport paths that not flow into one of the $n$ nodes of Image 2 penalize the cost by traversing the edges connected to $u_2$. From conservation of mass one can see that these transport paths satisfy $P^a_{j u_2} = H_{ja} - (1/n)\sum_i G_{ia}$, $\forall j \in V_2$. Thus, summing over $a$ and $j$ returns exactly $\sum_{aj} P_{j u_2}^a = ||\sum_j H_j - \sum_i G_i||_1$, with the $1$-norm taken over the commodities. This is precisely the penalization we introduced in \cref{apxeqn:relax}.

\end{itemize}

\section{Equivalence between multicommodity dynamics and OT setup}

With the following derivations (similar to \cite{lonardi2021designing, bonifaci2022physarum}), we show that asymptotic solutions of Eqs. (3)-(4) (main text) are equivalent to minimizers of Eq. (1) (main text). This implies that by solving the multicommodity dynamics we find a solution of the multicommodity OT minimization problem. More practically, for a given pair of images, running a numerical scheme on Eqs. (3)-(4) (main text) allows us to compute $\lim_{t \to \infty} P(t) = P^\star$, hence $J_\Gamma^\star = J_\Gamma |_{P = P^\star}$, and use the latter as a measure of similarity between them.

More in detail, we first demonstrate the equivalence between stationary solutions of the multicommodity dynamics and minimizers of the multicommodity OT problem introducing a second accessory minimization problem. Stationary solutions are proven to be asymptotes of Eq. (4) (main text) only afterwards, with the introduction of a Lyapunov functional for the multicommodity dynamics.

\subsection{Stationary solutions of the multicomodity dynamics and OT minimizers}

Initially, we observe that stationary solutions of the multicommodity dynamics satisfy the relation
\begin{equation}
\label{eqn:scaling_dyn}
 x_e = || P_e ||_2^{2/(1+\gamma)} \quad \forall e \in E,
\end{equation}
that one can derive setting the left hand side of Eq. (4) (main text) to zero, defining $P^a_e = x_e(\phi_i^a - \phi_j^a)/C_e$ for $e = (i,j)$, and $\gamma = 2-\beta$. We recover an scaling identical to \cref{eqn:scaling_dyn} introducing the following auxiliary constrained minimization problem:
\begin{align}
    \label{eqn:objunconstrained}
    &\min_{x,P} \left\{ \f{1}{2} \sum_e \f{C_e}{x_e}|| P_e ||_2^2 + \f{1}{2\gamma} \sum_e C_e x_e^\gamma \right\}\\
    \label{eqn:constraints}
    &\text{s.t. } \sum_e B_{ie} P_e^a = S_i^a \quad \forall i \in V, a = 1,\dots,M.
\end{align}
In fact, differentiating with respect to $x_e$ the objective function in \cref{eqn:objunconstrained}, and setting the derivatives to zero, yields
\begin{equation}
    \label{eqn:scaling}
    -\f{C_e}{x_e^2} || P_e ||_2^2 + C_e x_e^{\gamma-1} \overset{!}{=} 0 \quad \longrightarrow \quad x_e = || P_e ||_2^{2/(1+\gamma)} \quad \forall e \in E.
\end{equation}
Noticeably, \cref{eqn:objunconstrained} admits a straightforward physical interpretation. In fact, the first term $J = (1/2) \sum_e C_e||P_e||_2^2/x_e$ is Joule's first law. Particularly, transport paths can be thought of as fluxes of mass transported through the edges of a capacitated network with resistances $r_e = C_e / x_e$. While the second term, $W_\gamma = (1/2\gamma) \sum_e C_e x_e^\gamma$, is the cost needed to build the network infrastructure. The constraints in \cref{eqn:constraints}---identical to Eq. (3) (main text)---are equivalent to Kirchhoff's law, enforcing conservation of mass.

Most remarkably, the scaling of \cref{eqn:scaling} can be also recasted in \cref{eqn:objunconstrained} to find that $J_\Gamma = J + W_\gamma$ (neglecting multiplicative constants). This connects the multicommodity dynamics with the objective function of Eq. (1) (main text). In detail,
\begin{align}
    J + W_\gamma &= \f{1}{2} \sum_e \f{C_e}{x_e}|| P_e ||_2^2 + \f{1}{2\gamma} \sum_e C_e x_e^{\gamma} \\
    &\overset{\text{\cref{eqn:scaling}}}{=} \f{1}{2} \sum_e C_e|| P_e||_2^{2\gamma/(1+\gamma)}+ \f{1}{2\gamma} \sum_e C_e|| P_e||_2^{2\gamma/(1+\gamma)}\\
    &\overset{\Gamma = 2\gamma / (1+\gamma)}{=} \f{1}{\Gamma} \sum_e C_e|| P_e||_2^{\Gamma}\\
    &= \f{1}{\Gamma} J_\Gamma (G,H).
\end{align}

To complete the mapping between the multicommodity dynamics and the minization setup, we show that the space of transport tensors $\Pi(G,H)$ is exactly the same space defined by \cref{eqn:constraints}. This can be seen with the following chain of equalities:
\begin{alignat}{2}
    \label{eqn:chain1}
    \sum_k P^a_{ik} - \sum_j P^a_{ji} &= G_i^a - H_i^a \quad &&\forall i \in V, a = 1,\dots,M\\
    \label{eqn:chain2}
    \sum_k P^a_{e=(i,k)} - \sum_j P^a_{e=(j,i)} &= S_i^a \quad &&\forall i \in V, a = 1,\dots,M\\
    \label{eqn:chain3}
    \sum_e B_{ie} P^a_e &= S_i^a \quad &&\forall i \in V, a = 1,\dots,M.
\end{alignat}
Here we take the difference between the OT constraints of $\Pi(G,H)$ in \cref{eqn:chain1}, we then use the definition of $S$ in \cref{eqn:chain2}, and compact the plus and minus signs using the signed incidence matrix $B$ in \cref{eqn:chain3}. This allows us to recover Kirchhoff's law as formulated in \cref{eqn:constraints} and Eq. (3) (main text).

\subsection{Multicommodity dynamics asymptotes: Lyapunov functional}

We complete our discussion introducing the Lyapunov functional for Eq. (4) (main text) proposed in \cite{lonardi2021designing, bonifaci2022physarum}. The functional reads:
\begin{align}
    \mathcal{L}_\gamma[x] = \f{1}{2} \sum_{ai} \phi_i^a[x]S_i^a + \f{1}{2\gamma} \sum_e C_e x_e^\gamma,
\end{align}
and it is a multicommodity generalization of that originally introduced in \cite{bonifaci2012physarum}.
This is a well-defined Lyapunov functional for the multicommodity dynamics, in fact, along a curve $x(t)$ solution of Eq. (4) (main text),
\begin{align}
    \f{d \mathcal{L}_\gamma [x(t)] }{dt} \leq 0.
\end{align}
With the equality satisfied if and only if $x(t)$ is a stationary point of Eq. (4) (main text). This can be shown as follows. We claim that
\begin{align}
    \label{eqn:claim}
    \f{\partial \mathcal{L}_\gamma}{\partial x_e} = \f{C_e}{2} \left( x_e^{\gamma-1} - \f{|| \phi_i - \phi_j ||_2^2}{C_e^2} \right) \quad \forall e = (i,j) \in E.
\end{align}
This equality can be retrieved differentiating both sides of Eq. (3) (main text) by $x_e$, thus obtaining
\begin{align}
    \sum_j \frac{\partial L_{ij}}{\partial x_e} \phi_j^a + \sum_j L_{ij}\frac{\partial \phi_j^a}{\partial x_e} = 0 \quad \forall i \in V,e \in E,a = 1,\dots,M,\\
    \label{eqn:derivation}
    \sum_j L_{ij}\frac{\partial \phi_j^a}{\partial x_e} = - \sum_j B_{je} ({1}/{C_e}) B_{ie} \phi_j^a \quad \forall i \in V,e \in E,a = 1,\dots,M.
\end{align}
Then, multiplying \cref{eqn:derivation} by $\phi_i^a$ and summing over $i$ one gets
\begin{align}
    \sum_{ij} \phi_i^a L_{ij}\frac{\partial \phi_j^a}{\partial x_e} &= - \sum_{ij} \phi_i^a B_{ie} ({1}/{C_e}) B_{je} \phi_j^a \quad \forall e \in E,a = 1,\dots,M,
\end{align}
further summing over $a$ yields
\begin{align}    
    \label{eqn:derivation2}
    \f{\partial}{\partial x_e} \left( \sum_{aj} S_j^a \phi_j^a \right) &= - C_e \frac{|| \phi_i - \phi_j ||_2^2}{C_e^2} \quad \forall e = (i,j) \in E,
\end{align}
where in the left hand side of \cref{eqn:derivation2} we used Eq. (3) (main text). From \cref{eqn:derivation2} the equality in \cref{eqn:claim} follows immediately. Now, thanks to \cref{eqn:claim} we can prove that the Lie derivative of the functional is less than or equal to zero. In fact,
\begin{align}
    \f{d \mathcal{L}_\gamma}{dt} &= \sum_e \frac{\partial \mathcal{L}_\gamma }{ \partial x_e} \frac{d x_e}{dt}\\
    &\overset{\text{\cref{eqn:claim}}}{=} \sum_e \f{C_e}{2} \left( x_e^{\gamma-1} - \f{|| \phi_i - \phi_j ||_2^2}{C_e^2} \right) \frac{d x_e}{dt}\\
    \label{eqn:lessthanlyap}
    &\overset{\text{Eq. (4)}, \gamma = 2-\beta}{=} - \sum_e \f{C_e}{2} x_e^{2-\gamma} \left( x_e^{\gamma-1} - \f{|| \phi_i - \phi_j ||_2^2}{C_e^2} \right)^2 \leq 0.
\end{align}
With the equality in \cref{eqn:lessthanlyap} that is recovered if and only if (i) $x_e = 0$, or (ii) the scaling in \cref{eqn:scaling} holds.

Finally, we show that the Lyapunov is identical to the total sum of dissipation and transport cost, i.e., $\mathcal{L}_\gamma = J + W_\gamma$. This can be done multiplying both sides of Eq. (3) (main text) by $\phi_i^a$ and then summing over $i$ and $a$, namely
\begin{align}
    \sum_{aiej} \phi_i^a B_{ie} (x_e/C_e) B_{je} \phi_j^a = \sum_{ai} \phi_i^a S_i^a\\ 
    \sum_{e} \f{C_e}{x_e} || P_e ||_2^2 = \sum_{ai} \phi_i^a S_i^a 
\end{align}
where we used $P^a_e = x_e(\phi_i^a - \phi_j^a)/C_e$, for $e = (i,j)$. This allows us to conclude.

In summary, we showed that the multicommodity dynamics admits a well-defined Lyapunov functional, which is equivalent to the sum of a dissipation and an infrastructure cost. These two contributions, which are jointly minimized by Eq. (4) (main text), when evaluated along their minimizers correspond to the multicommodity OT cost $J_\Gamma$ of Eq. (1) (main text). Introducing the Lyapunov functional is crucial to formally show that asymptotics of the dynamics are equivalent to minimizers of the cost, namely $\lim_{t\to\infty} P(t) = P^\star$.

Lastly, we remark the effect of $\gamma$ (resp. $\beta$) on the minimization problem. In the setting where $\gamma > 1$ ($\beta < 1$) the functional $\mathcal{L}_\gamma$ is convex, with one unique minimizer. For $\gamma < 1$ ($\beta > 1$) the functional landscape becomes rugged and strongly non-convex, with multiple minimizers each correspondent to a local minima of the cost. Hence, in this second scenario, running Eq. (4) (main text) permits to converge in a stationary point, which however may not be its global minimum.

\section{Cross-Validation: Flowers Dataset}\label{sec:cv_flowers} 

We perform a 4-fold cross validation on both parameters used for the construction of the ground cost, $\theta$ and $\tau$, and on algorithms' regularization parameters, $\beta$ and $\varepsilon$. We briefly summarize it in this section.

The JF30 Dataset \cite{seeland2017plant} is made of 1,479  elements, divided in 30 classes. First, we separate it into two subsets: \emph{train} and \emph{test},  with classes' frequencies being the same in these subsets as in the entire dataset.  To cross-validate our methods, we further separate the train set into 4 folds of equal size, each to be used in turn as a validation set. More in detail, each experiment is executed fixing the validation fold and an image belonging to it, then, the Optimal Transport costs $J_\Gamma^\star$ between such image all the other images in the train set---made of the other three folds---is calculated. This procedure is repeated for all images in the validation set, and swapping each of the 4 train folds as validation set.  We use a $k$-nearest neighbors classifier over $J_\Gamma^\star$ to assign to an image in the validation set its label, that is, for each validation image we consider the $k$ train samples with lowest $J_\Gamma^\star$, and label the validation sample with the most frequent class among these $k$. This allows us to calculate the classification accuracy of a given fold, and then to average the accuracy over the 4 permutations of the validation and train set. The total amount of experiments we ran in order to cross-validate the model is approximately 50,000.

\begin{figure}[b] %
    \centering
      \includegraphics[width=\textwidth]{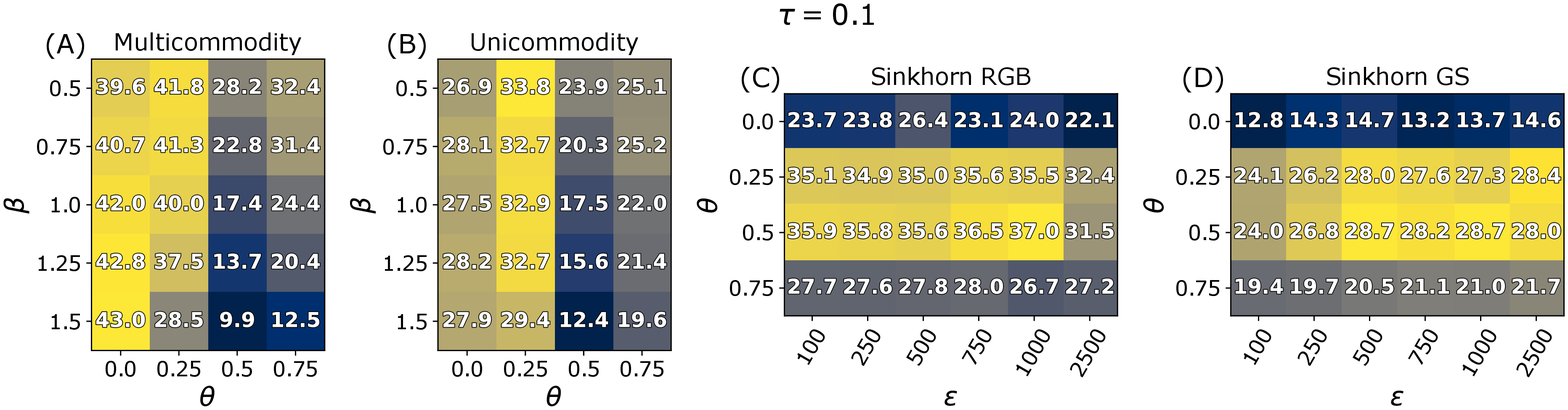}
    \caption{Cross-validation results for $\tau=0.1$. Figures are accuracy values obtained with the $4$-fold cross validation on JF30. Cells are colored with a darkest-to-brightest palette based on the accuracies. Subplots correspond to: (\textbf{A}) the multicommodity dynamics, (\textbf{B}) the unicommodity dynamics, (\textbf{C}) Sinkhorn on colored images, and (\textbf{D}) Sinkhorn on grayscale images.}
    \label{apxfig:cv_01}
\end{figure}

\begin{figure}[htpb] 
    \centering
          \includegraphics[width=\textwidth]{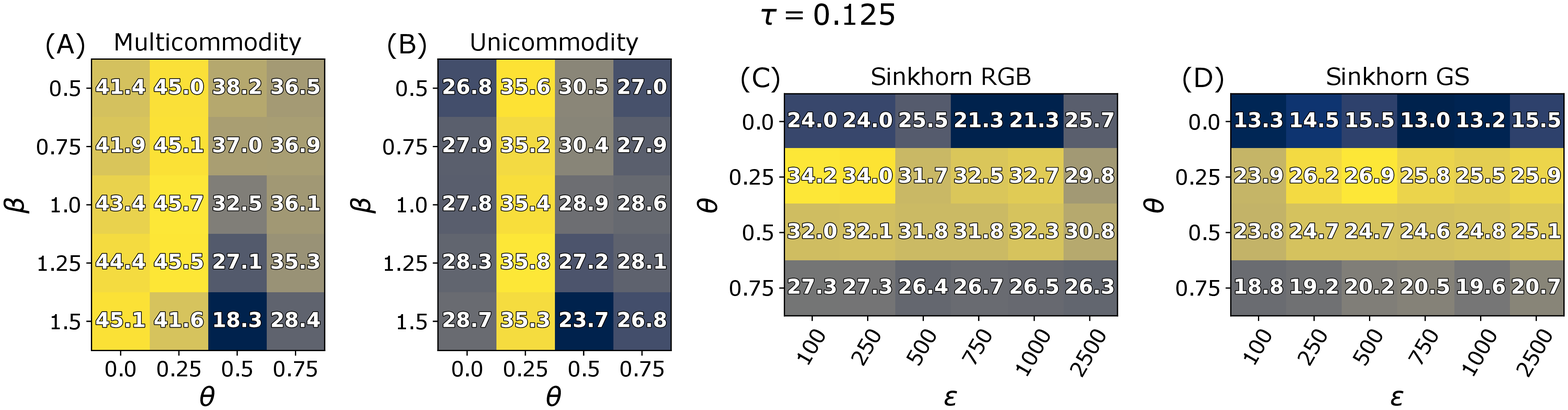}
    \caption{Cross-validation results for $\tau=0.125$. Figures are accuracy values obtained with the $4$-fold cross validation on JF30. Cells are colored with a darkest-to-brightest palette based on the accuracies. Subplots correspond to: (\textbf{A}) the multicommodity dynamics, (\textbf{B}) the unicommodity dynamics, (\textbf{C}) Sinkhorn on colored images, and (\textbf{D}) Sinkhorn on grayscale images.}
    \label{apxfig:cv_0125}
\end{figure}

Results are shown in \Cref{apxfig:cv_01} and \Cref{apxfig:cv_0125}. These depict the average accuracy of: (A) the multicommodity $(M=3)$ dynamics; (B) the unicommodity $(M=1)$ dynamics, both for $\beta \in \{0.5,0.75,1,1.25,1.5\}$; (C) Sinkhorn algorithm on colored images (Sinkhorn RGB); and (D) Sinkhorn algorithm on grayscale images (Sinkhorn GS), for $\varepsilon \in \{100, 250, 500, 750, 1000, 2500\}$. Letters in parentheses refer to those of \Cref{apxfig:cv_01} and \Cref{apxfig:cv_0125}. The regularization parameters are validated together with $\tau \in \{0.1,0.125\}$ and $\theta \in \{0, 0.25, 0.5, 0.75 \}$. Both multicommodity and unicommodity dynamics are have initial conditions $x_e(0)=1, \forall e \in E$.

All figures displayed in \Cref{apxfig:cv_01} and \Cref{apxfig:cv_0125} correspond to highest accuracies returned by the $k$-NN classifier, with $k = 1,2,\dots,20$. Observing the results, one can see that best performances are attained at $(\tau, \theta, \beta) = (0.125,0.25,1)$ for the multicommodity dynamics, and at $(\tau, \theta,  \beta) = (0.125, 0.25, 1.25)$ for the unicommodity dynamics.

Noticeably, the accuracy monotonically increases (resp. decreases) with $\beta$ for a fixed value of $\theta$, namely $\theta = 0$ (resp. $\theta=0.25$). This can be addressed to the fact that, when no color information is taken into account in the construction of the ground metric ($\theta=0$), it is more advantageous to consolidate transport paths on cheap edges correspondent to pixels whose positions are close, thus choosing a larger $\beta$. On the other hand, introducing colors in $C$ ($\theta>0$), and thus creating a more disordered ground cost matrix, favors distributing transport paths on the network (See Model Interpretability Section in this SM).

Remarkably, $\tau$ also has an impact on the classification accuracy of our algorithms: the larger we set its value to be---thus trimming less edges from the transport network---the more accurate the classification becomes. This behavior is evidently different for Sinkhorn algorithm, as explained here below.


\begin{figure}[b]
    \centering
          \includegraphics[width=0.9\textwidth]{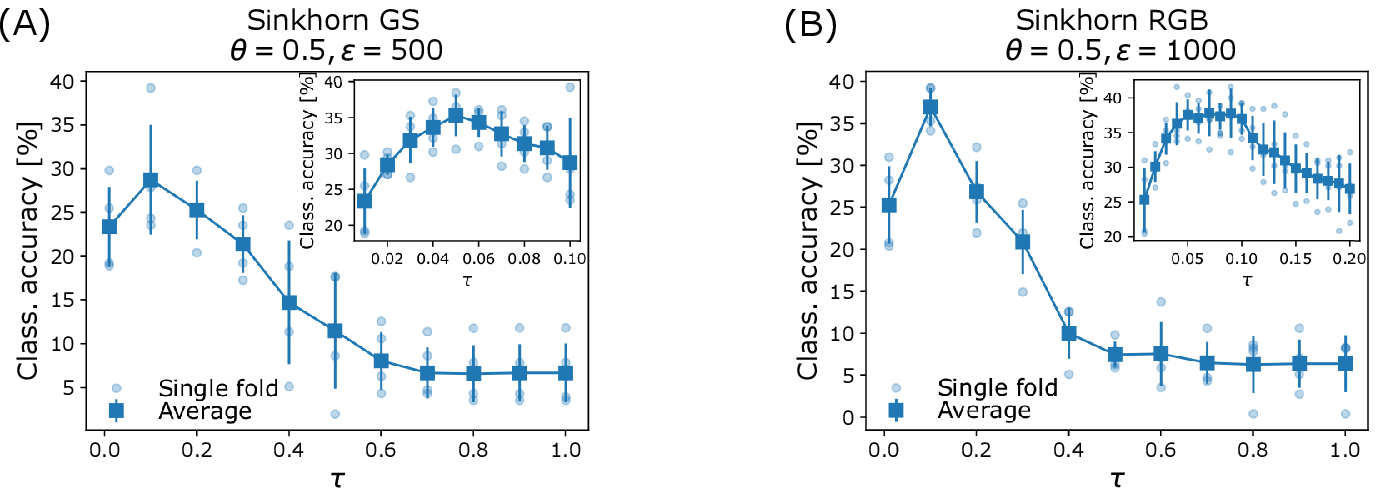}
    \caption{Sinkohrn's cross-validation results varying $\tau$. Subplots correspond to: (\textbf{A}) Sinkhorn GS, (\textbf{B}) Sinhorn RGB. In each subplot, circular markers correspond to the accuracy values of each fold, instead squares and bars represent their average and standard deviations. In the insets, we refined the grid of $\tau$ in an interval of interest, where classification accuracy is peaked.}
    \label{apxfig:cv_all}
\end{figure}

Cross-validation of Sinkhorn algorithm is taken a step further. Motivated by the classification accuracy drop observed in \Cref{apxfig:cv_01}, \Cref{apxfig:cv_0125} [(C), (D)] when enlarging the trimming threshold from $\tau = 0.1$ to $\tau = 0.125$, we fix $\theta$ and $\varepsilon$ to the best values in \Cref{apxfig:cv_01} [(C), (D)], and progressively reduce $\tau$. Results are shown in \Cref{apxfig:cv_all} (A) for Sinkhorn on grayscale images, and in \Cref{apxfig:cv_all} (B) for Sinkhorn on colored images.

Notice that both Sinkhorn GS and Sinkhorn RGB returns bell-shaped curves when changing $\tau$. In particular, low classification accuracy is attained when strongly reducing $\tau$, as well as when the trimming threshold is high (approximately $\tau \geq 0.5$). In the first case, many elements of the ground cost matrix are cut, and not enough information is taken into account into the OT setup to properly perform classification. In the second, too much noise is included in into $C$, which also negatively affects classification.
More in detail, we observe in \Cref{apxfig:cv_all} (a, inset), that Sinkhorn GS performs best when $\tau = 0.05$. For Sinkhorn RGB, i.e. \Cref{apxfig:cv_all} (b, inset), there is a plateau for all values of the threshold within the interval $[0.05,0.1]$.

\begin{figure}[b]
\centering
\includegraphics[width=0.5\linewidth]{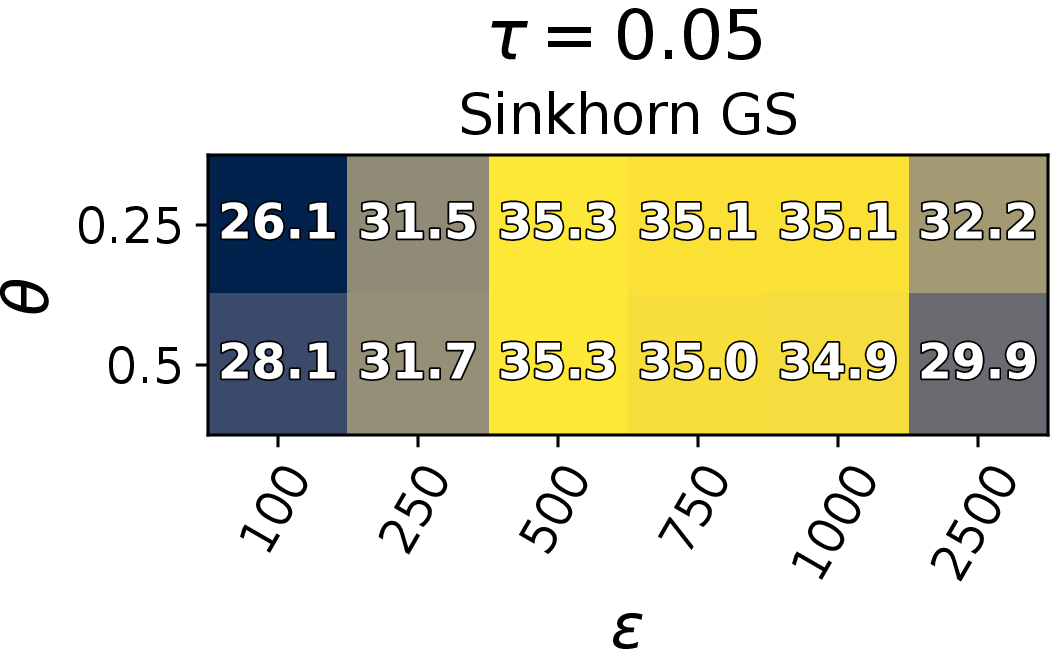}
\caption{Refined cross-validation results Sinkhorn GS and $\tau=0.05$. Figures are accuracy values obtained with the $4$-fold cross validation on JF30. Cells are colored with a darkest-to-brightest palette based on the accuracies.}
\label{apxfig:cv_005}
\end{figure}

These observations lead us to the choice of $\tau = 0.05$ for Sinkhorn GS, that we re-cross-validate ranging $\theta \in \{ 0.25, 0.5 \}$ and $\varepsilon \in \{100,250,500,750,1000,2500\}$. Looking at the results in \Cref{apxfig:cv_005}, we note that optimal parameters for Sinkhorn GS are $(\theta, \varepsilon) = (0.25, 500)$ and $(\theta, \varepsilon) = (0.5, 500)$, which return identical classification accuracy.

As for Sinkhorn RGB, we fix the trimming threshold at the two ends of the plateau in \Cref{apxfig:cv_all} (b, inset), $\tau = 0.05$ and $\tau = 0.1$, and re-cross-validate them with $(\theta, \varepsilon) = (0.25, 100)$ and $(\theta, \varepsilon) = (0.5, 1000)$. Here, we choose two disparate values of $\theta$ and $\varepsilon$ not being able to observe a clear relation between these two variables in \Cref{apxfig:cv_01} (C) and \Cref{apxfig:cv_0125} (C). Namely, $\varepsilon = 100$ (low) and $\theta = 0.25$ perform better for $\tau = 0.125$, in contrast to $\varepsilon = 1000$ (high) and $\theta = 0.5$ for $\tau = 0.1$. Results are in \Cref{apxtable:sinhornRGB}, optimal parameters are $(\theta, \tau, \varepsilon) = (0.25, 0.05, 100)$.

\begin{table}[h]
    \centering
    \begin{tabular}{lcccccc} \toprule
    Algorithm & \multicolumn{4}{c}{Hyperparameters} & {Class. accuracy} \\
    &  $\theta$ & $\tau$  & $\varepsilon$ & $k$ & [\%] ($\uparrow$)\\
    \midrule
    \multirow{4}{*}{\rotatebox[origin=c]{-35}{Sinkhorn RGB}} & 0.25 & 0.05 & 100 & 1 & 58.4\\
     & 0.5 & 0.05 & 1000  & 1 & 53.6 \\
     & 0.25 & 0.1 & 100 & 1 & 53.2\\
     & 0.5 & 0.1 & 1000 & 1 & 49.0\\
    \bottomrule
    \end{tabular}
{\caption{\normalsize{Refined cross-validation results for Sinkhorn RGB. Rows are sorted (from bottom to top) using the average percent accuracy obtained with the $4$-fold validation (from worst to to best). \label{apxtable:sinhornRGB}}}}
\end{table}

\section{Experimental details: Fruits Dataset}

Here, we describe in detail the experimental setup designed for the Fruit Dataset (FD) \cite{alves2018handwritten}. FD consists of 163 images of 15 fruit types. We split the whole dataset into \emph{train} and \emph{test} sets, each with 70\% and 30\% of the available images, respectively. As for the other dataset, classes' frequencies are the same in these subsets as in the entire dataset. Given the rather small size of this dataset, we directly perform classification comparing train and test. All the experiments have been executed with the two best performing parameter configurations of  $\varepsilon$ and $\theta$, cross-validated on JF30, for Sinkhorn-based methods. These are: $(\theta,\varepsilon)=(0.25,500)$, $(\theta,\varepsilon)=(0.5,500)$ for Sinkhorn GS [see \Cref{apxfig:cv_01} (D)], and $(\theta,\varepsilon)=(0.5,1000)$, $(\theta,\varepsilon)=(0.5,750)$ for Sinkhorn RGB [see \Cref{apxfig:cv_01} (C)]. For our dynamics, we selected the two best performing values of $\beta$, for $\theta = 0$ and $\theta = 0.25$. Namely, $(\theta,\beta)=(0.25,1)$, $(\theta,\beta)=(0.25,1.5)$ for the multicommodity dynamics [see \Cref{apxfig:cv_0125} (A)], and $(\theta,\beta)=(0.25,1.25)$, $(\theta,\beta)=(0,1.5)$ for the unicommodity dynamics [see \Cref{apxfig:cv_0125} (B)]. The trimming threshold is ranged in $\tau \in \{0.04, 0.05, 0.06, 0.07\}$.

\section{Image preprocessing}

The elements of both datasets are processed in the following way. First, each image is coarsened with an average pooling, the only input needed for this step is the size of the square mask, $\texttt{ms}$. Its stride is in fact set to $\texttt{stride} = \texttt{ms}$, and the padding to $\texttt{pad} = 0$. All images were conveniently trimmed so that both their widths and heights are divisible by the pooling mask size. We set $\texttt{ms} = 40$ for JF30, and $\texttt{ms} = 30$ for FD. Furthermore, we smooth the images using a Gaussian filter on each color channel, with standard deviation $\sigma = 0.5$. 

Moreover, to convert colored images into grayscale ones, which are given as input to Sinkhorn GS and to our unicommodity dynamics $(M=1)$, we preproces them as follows. Let $(R,G,B)$, be the three color channels composing each pixel of a colored image, these are converted into a unique channel (its grayscale counterpart), whose intensity $I$ is calculated with the weighted sum $I = 0.2125 R + 0.7154 G + 0.0721 B$. The weights correspond to those used by cathode-ray tube (CRT) phosphors as they are more suitable to represent human perception of red, green and blue than equally valued weights \cite{skimage}.

\subsection{Color distributions of images}

\begin{figure}[t]
    \centering
    \includegraphics[width=\linewidth]{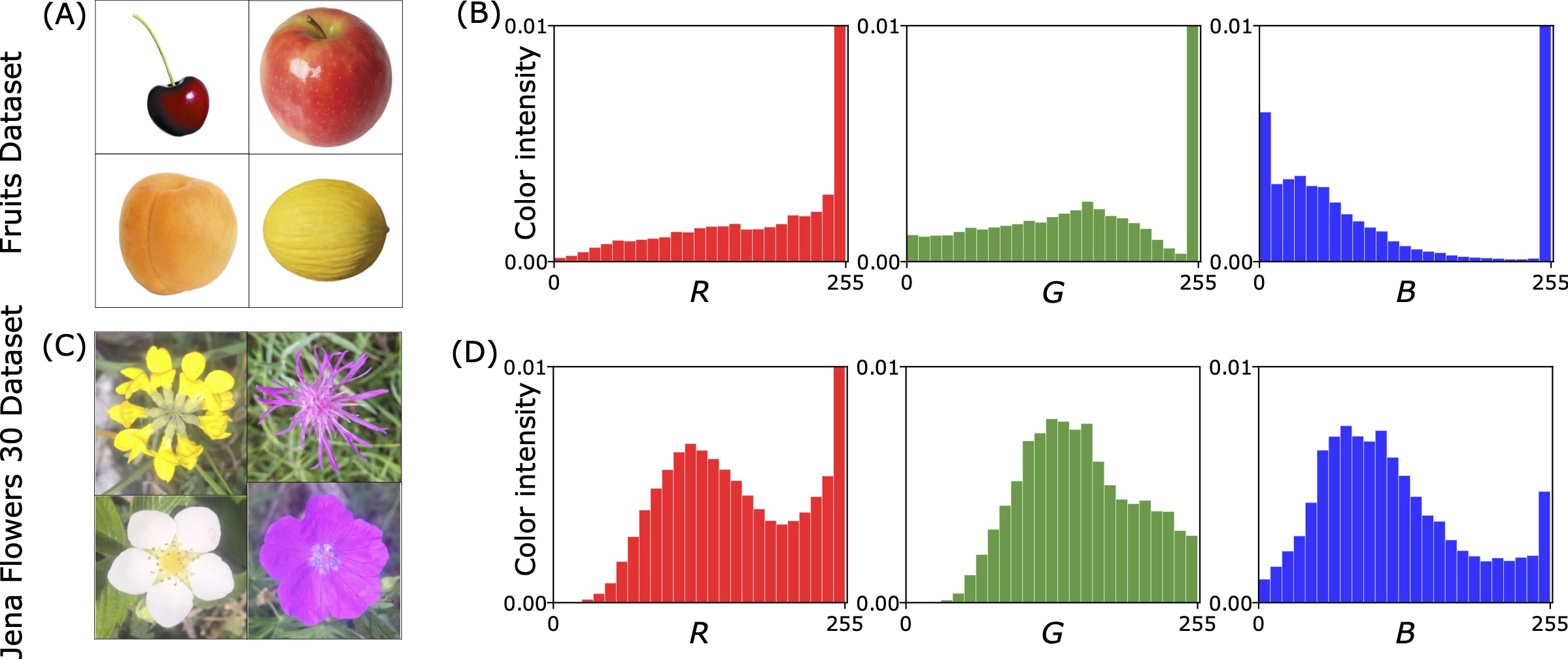}
    \caption{Color distributions in the two datasets. Subplots (\textbf{A}), (\textbf{B}) are relative to FD, subplots (\textbf{C}), (\textbf{D}) to JF30. In (A), (C) we plot four random images drawn from the two datasets. In (B), (D) the average color intensities (properly normalized to sum to one) of 100 random samples extracted from the two datasets.  The plots correspond to red $= R$, green $= G$, and blue $= B$.\label{apxfig:color_dist}}
\end{figure}

As shown in Table I (main text), for the multicommodity and the unicommodity dynamics, optimal values of the trimming threshold $\tau$ are much lower in the Fruit Dataset \cite{alves2018handwritten} than in the Jena Flowers 30 Dataset \cite{seeland2017plant}. This can be addressed to the fact that color distributions of fruits, belonging to the first dataset, are drastically light-tailed compared to those of flowers in the second dataset.  Thus, the cost $C$ is naturally noisier in the latter case, and a larger trimming is necessary to remove such noise from classification. 

Most of the noise in pictures of flowers comes from the background. In fact, while all flowers are photographed in nature, fruits are depicted on a white background. This can be seen in \Cref{apxfig:color_dist} (A)-(D). In subplot (A) we show four images randomly sampled from the Fruit Dataset, in (C) four random samples of the Jena Flowers 30 Dataset.  In (B) and (D) we plot the average color intensity of the RGB channels for 100 random samples belonging to the two datasets.  Here,  the histograms in (B) are relative to the fruits, those in (D) to the flowers. From the plots it can be clearly seen that the color distributions of  \Cref{apxfig:color_dist} (B) are starkly peaked around $(R,G,B) = (255,255,255) =$ white in standard RGB encoding.

\section{Sinkhorn benchmarks} 

In our experiments, we compare the multicommodity and unicommodity dynamics against Sinkhorn algorithm, popularized by the seminal work of \cite{cuturisinkhorn}. The idea of Sinkhorn is to regularize the standard OT problem by adding an entropic barrier to the cost function. More in detail, and following the notation adopted in our manuscript, the minimization problem proposed in \cite{cuturisinkhorn} is:
\begin{align}
\label{eqn:sinkhorn_apx}
\min_{\substack{P \, \text{s.t.} \, \sum_j  P_{ij} = g_i \\ \qquad\, \sum_i  P_{ij} = h_j }} \left\{ \sum_{ij} P_{ij}C_{ij} - \varepsilon h(P) \right\}, \qquad {h(P) = -\sum_{ij} P_{ij} \log P_{ij}}.
\end{align}
Here transport paths $P$, which generally lie in the polyhedral set described by the constraints $\sum_j P_{ij} = g_i\, \forall i$ and $\sum_i P_{ij} = h_j \, \forall j$, are smoothed by the entropy $h(P)$. This trick makes the optimization problem strictly convex, and permits to solve it with a very efficient matrix scaling algorithm---Sinkhorn's fixed point iteration.

We generalize the problem in \cref{eqn:sinkhorn_apx} in order to take in account transport tensors, $G$ and $H$, which carry information of multiple color channels, and transport paths $P$. In detail, we propose the following minimization problem for each commodity---color channel---$a$,
\begin{align}
\label{eqn:sinkhorn_colors}
\min_{\substack{P^a \, \text{s.t.} \, \sum_j  P^a_{ij} = G^a_i \\ \qquad \;\; \,\, \sum_i P^a_{ij} = H^a_j }} \left\{ J_\text{sink}^a = \sum_{ij} P^a_{ij}C_{ij} - \varepsilon h(P^a) \right\}, \qquad {h(P^a) = -\sum_{ij} P^a_{ij} \log P^a_{ij}}.
\end{align}
This allows to efficiently compute, using Sinkhorn's scaling, an Optimal Transport path $P^a_\text{opt}$ for each commodity, together with its correspondent optimal cost $J_\text{sink,opt}^a = J_\text{sink}^a |_{P^a = P^a_{\text{opt}}}$. Finally, the Optimal Transport cost for colored images is calculated as $J^\text{RGB}_\text{sink,opt} = (1/3) \sum_{a=1}^{M=3} J_\text{sink}^a$.

\section{Model interpretability}

In this section we discuss the effect that the parameters $\theta$, $\tau$, and $\beta$ have one the OT setup.

\begin{figure}[htpb]
    \centering
    \includegraphics[width=\linewidth]{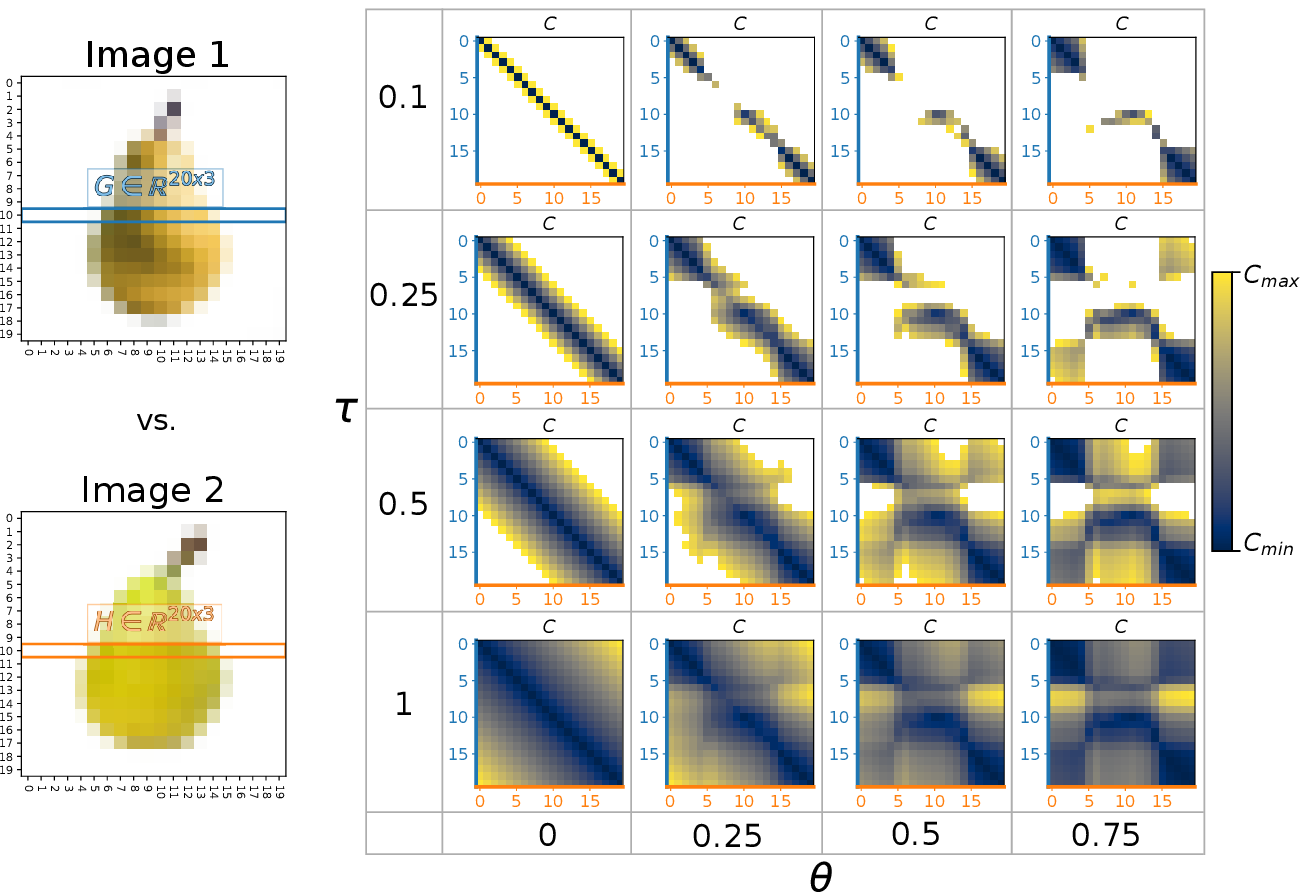}
    \caption{Effect of $\theta$ and $\tau$ on OT. On the left, we display the two samples used to build the ground costs $C$. Highlighted rows in blue and orange are those considered to extract $G$ and $H$. On the right side of the panel we plot $C$ for $\theta = \{ 0, 0.25, 0.5, 0.75\}$ and $\tau = \{ 0.1, 0.25, 0.5, 1\}$. White regions correspond to trimmed values, i.e. entries of $C$ that are larger than $\tau$.}
    \label{apxfig:model_interpretability_1}
\end{figure}

First, we explain the experiment in \Cref{apxfig:model_interpretability_1}. We start by sampling two images of the FD dataset belonging to the same class. These images have identical shape, i.e. width $w$ and height $h$ equal to 20. They are displayed on the leftmost part of \Cref{apxfig:model_interpretability_1}. From these two images, we obtain the tensors $G$ and $H$, that are transported in the OT problem. The first, $G$, is constructed using all the pixels on the $11$th row of Image 1, thus its dimension is $m \times M = 20 \times 3$. The same row of Image 2 is used to build $H$, also in this case its size is $n \times M = 20 \times 3$.

The two tensors enter in Eq. (1) (main text) together with a $(20\times20)$-dimensional cost $C$, which is built with pixels' locations and color information using \Crefrange{apxeqn:C}{apxeqn:X}. The scope of this discussion is to refine the intuition on these formulas, and on the effect that $\theta$ and $\tau$ have on $C$. In \Cref{apxfig:model_interpretability_1} we plot the ground cost $C$ for the two tensors $G$ and $H$, for $\theta = \{ 0, 0.25, 0.5, 0.75\}$ and $\tau = \{ 0.1, 0.25, 0.5, 1\}$. All entries $C_{ij} > \tau$---which correspond to those edges that are trimmed from the transport network---are colored in white.

Notice that for $\theta = 0$ all costs are symmetric.  Indeed in this case  $C_{ij} = \min \{Y_{ij}, \tau\}$, with $Y$ that is containing the Euclidean distances between pixels' coordinates, i.e. \Cref{apxeqn:V}. Here, decreasing the trimming threshold $\tau$ progressively sparsifies the banded matrices drawn in the first column of \Cref{apxfig:model_interpretability_1}, with limit cases being $C = \text{diag}[C_{0,0}, C_{1,1}, \dots ,C_{19,19}]$---for $\tau$ sufficiently small, and $C = Y$---for $\tau \geq \max_{ij}{Y_{ij}}$. On the other hand, the symmetry is gradually broken as $\theta$ is increased, namely, when colors of the images are are used to build into $C$. This is clearly depicted in \Cref{apxfig:model_interpretability_1}, where the heatmaps get progressively disordered for larger values of $\theta$ (from left to right).

\begin{figure}[t]
    \centering
    \includegraphics[width=\linewidth]{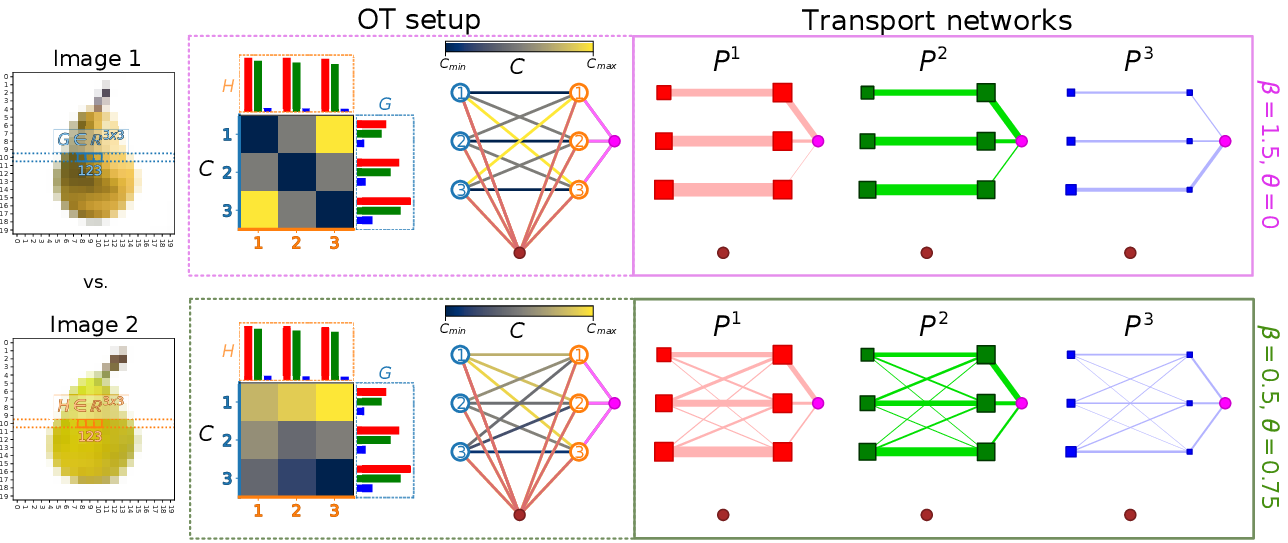}
    \caption{Effect of $\beta$ on OT. In the leftmost portion of the panel we plot Image 1 and Image 2, used in the OT problem. From these we extract the $(3\times3)$-dimensional tensors $G$ and $H$. These are drawn together with a heatmap of the cost $C$, and with the correspondent transport network. Color scales of edges and of entries of $C$ are identical. We also use the same numbering and color scheme for tensors' entries, indexes of $C$, and network nodes. Brown and magenta auxiliary nodes and edges are added after trimming, and after relaxing Kirchhoff's law. On the right side of the panel we plot the transport network again, but with edge thickness proportional to the Optimal Transport paths retrieved from Eqs. (3)-(4) (main text), and with colors correspondent to those of the commodities $a$. Node sizes are proportional to the values of $g^a$ and $h^a$, for $a=1,\dots,3$.}
    \label{apxfig:model_interpretability_2}
\end{figure}

To further expand this discussion, we design a second experiment, schematically represented in \Cref{apxfig:model_interpretability_2}. Here, we solve the OT problem between two tensors, $G$ and $H$, built similarly to those of \Cref{apxfig:model_interpretability_1}. Particularly, we consider three central pixels of the $11$th rows of Image 1 and Image 2, as drawn in the leftmost part of the Figure, so that both $g^a$ and $h^a$ are $3$-dimensional arrays for all $ a = 1,\dots,3$, and $C$ is a $(3\times3)$-dimensional matrix.

Depending on the values of $\theta$, the ground cost $C$ is either symmetric ($\theta = 0$), and computed only using pixels' coordinates, or strongly irregular ($\theta=0.75$), since colors of images are taken into account. In the first case, the transport network connecting the images has also a symmetric structure. Here, elements along the diagonal of the cost---correspondent to horizontal edges connecting orange and blue nodes with the same index---are much cheaper than all the other entries. This is due to the fact that the Euclidean distance between two pixels with the same position is zero (practically set to a safety default value $\epsilon = 10^{-5}$). Conversely, in the second case, introducing colors in the ground cost translates into having higher values along the diagonal elements of $C$. Here, colors---which distribute more smoothly on images---smooth out the cost as well, whose entries are more homogeneous.

As shown in \Cref{apxfig:cv_01} and \Cref{apxfig:cv_0125}, taking a purely Euclidean ground cost $C$, i.e., $\theta = 0$, returns higher classification accuracy when $\beta = 1.5$. Instead, building $C$ mostly with color information, thus setting $\theta = 0.75$, favors $\beta = 0.5$. We address this tendency to the effect that $\beta$ has on transport paths' consolidation, and we represent it on the rightmost portion of \Cref{apxfig:model_interpretability_2}, where we plot the Optimal Transport paths $\{ {P}^1, {P}^2, {P}^3\}$ obtained running Eqs. (3)-(4) (main text) on the OT setup just discussed.  In detail, for $\theta = 0$,  horizontal edges in the transport network are much cheaper than the others, therefore strong consolidation of transport paths ($\beta = 1.5$) benefits classification. Conversely, since for $\theta = 0.75$ the entries of $C$ are more homogeneous, distributing transport paths ($\beta = 0.75$) naturally reflects the topology of the transport network and allows to achieve better classification performances.

Lastly, we remark that transport paths do not pass through any transshipment edge (colored in brown in \Cref{apxfig:model_interpretability_2}) since $\tau$ is conveniently set to be sufficiently large. The auxiliary edges for Kirchhoff's law relaxation (colored in magenta) are instead traversed by transport paths since $G$ and $H$ are not normalized.

\section{Computational cost}

\subsection{Analytical discussion}

Considerable effort has been spent to reduce the high complexity burden of OT problems. The $O(|V|^2/\varepsilon^3)$ baseline of Sinkhorn algorithm \cite{sinkhorn1964relationship, knopp_sinkhorn, cuturisinkhorn}, where $|V|$ is the size of the histograms transported and $\varepsilon$ the parameter enforcing entropic regularization, is constantly improved. Notable recent results are the class of stochastic optimization algorithms proposed in \cite{genevay2016stochastic}, that have been ameliorated using greedy alternatives \cite{altschuler2017near} to achieve $\varepsilon$-approximation of the $1$-Wasserstein distance between two probability distributions in $O(|V|^2/\varepsilon^2)$ arithmetic operations \cite{lin2019efficient}.  Recently, an Adaptive Primal-Dual Accelerated Gradient Descent (APDAMD) scheme with complexity $O(\min \{ |V|^{9 / 4}/\varepsilon, |V|^{2}/\varepsilon^{2} \})$ for the same $\varepsilon$-perturbed problem has been presented in \cite{dvurechensky2018computational}.

In principle, our multicommodity method has a computational complexity of order $O(M|V|^2)$ for complete transport graph topologies, i.e., when edges in the transport network $K$ are assigned to all pixels' pairs. Nonetheless, we achieve a substantial decrease in complexity by sparsifying the graph with the trimming procedure of \cite{pele2008linear,pele2009fast}. Similarly to \cite{pele2008linear}, the final complexity of our algorithm is $O(M|V|)$. This improvement can be formally justified as follows, we start from a complete bipartite graph with $|E| = |V|^2/4$ (for simplicity $m = n = |V|/2$ is assumed). First, we trim expensive links, and reduce the number of edges of the transport network to $\langle K \rangle |V| + |V|$, where $\langle K \rangle$ is the average number of edges connected to a node that are not trimmed by $\tau$, and the second term $|V|$ counts the number of inflowing and outflowing transshipment links. Second,  we add $|V|/2$ links to the transport network to enforce Kirchhoff's law penalization, so that the final number of links amounts to $|E| = |V|(\langle K \rangle + 3/2)$, which is linear with respect to the number of nodes. 

Additionally, it is shown \cite{facca2020fast} that for $\beta = 1$, the Optimal Transport paths of the unicommodity OT problem on sparse topologies can be recovered with $z$ time steps as in Eq. (3) (main text), with $O(1) < z < O(|E|^{0.36})$. This bound has been found using a backward Euler scheme combined with the inexact Newton-Raphson method for the update of $x$, and solving Kirchhoff's law using an algebraic multigrid method.

\subsection{Experimental runtimes benchmarking against Sinkhorn}

We compare the runtime performances of the multicommodity dynamics of Eqs. (3)-(4) (main text), against the regularized Sinkhorn algorithm of \cite{schmitzer2019stabilized, chizat2018scaling}, implemented in POT: Python Optimal Transport \cite{flamary2021pot}, and for which we set the convergence threshold to $\widetilde{\varepsilon}_\text{sink} = 0.01$. Our implementation uses a forward Euler scheme for the discretization of Eq. (4) (main text), and a sparse direct linear solver (UMFPACK) for Eq. (3) (main text). Our code was run until convergence, achieved if $(J_\Gamma({n+1})-J_\Gamma({n}))/\Delta t < \widetilde{\varepsilon}_\text{dyn}$, i.e. when the relative cost difference evaluated at two consecutive iteration is below $\widetilde{\varepsilon}_\text{dyn} = 1$. We set the discretization time step $\Delta t = 0.5$.

All codes are executed on 20 pairs of images, randomly sampled from the Jena Flowers 30 Dataset \cite{seeland2017plant} and the Fruit Dataset \cite{alves2018handwritten}. We compare our multicommodity dynamics ($M=3$) against Sinkhorn algorithm on colored images, and the unicommodity dynamics ($M=1$) against Sinkhorn on grayscale images.
 
\begin{figure}[t]
    \centering
    \includegraphics[width=0.9\linewidth]{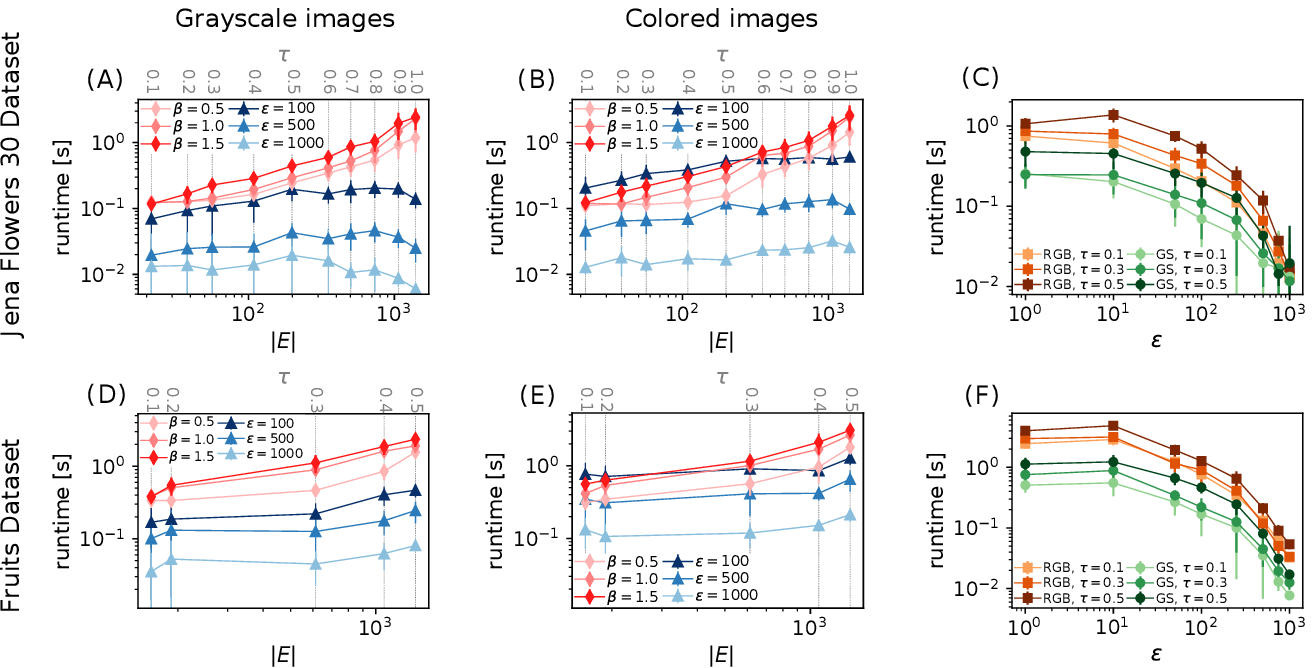}
    \caption{Runtimes of algorithms. Subplots (\textbf{A})-(\textbf{C}) are experiments on JF30, subplots (\textbf{D})-(\textbf{F}) are those on FD. In (A), (B), (C), and (D) we plot with red diamonds runtimes of our dynamics, with $M=1$ in (A), (D) and $M=3$ in (B), (E). Blue triangles are denote runtimes of Sinkhorn. Color shades correspond to different values of the regularization paramters. In (C) and (F) we show runtimes of Sinkhorn against $\varepsilon$, with orange and green markers used for colored and grayscale images, respectively. Color shades here denote different values of the trimming threshold $\tau$. Errorbars are standard deviations obtained over $20$ random image pairs.}
    \label{apxfig:runtime}
\end{figure}

 Results are shown in \Cref{apxfig:runtime}. Here, we plot elapsed times for the experiments on JF30 and on FD in the panels (A)-(C) and (D)-(F), respectively. Subplots from left to right represent runtimes for the algorithms executed on grayscale images [(A), (D)], on colored images [(B), (E)], and for Sinkhorn executed in both setups [(C), (F)].
 
 Observing \Cref{apxfig:runtime} [(A), (D)], we notice that runtimes for the multicommodity dynamics are larger than Sinkhorn's. Our algorithm converges faster if $\beta < 1$, i.e., when the multicommodity transport cost is convex. Setting $\beta > 1$ negatively affects convergence times. In general, for all values of $\beta$, increasing the trimming threshold $\tau$, and thus the average number of edges in the transport networks, leads to slower convergence. Sinkhorn algorithm is not as dependent on $|E|$, e.g., in \Cref{apxfig:runtime} (A), runtimes are approximately constant. Moreover, coherently to what expected \cite{cuturisinkhorn}, increasing the effect of the entropic barrier---enlarging $\varepsilon$---makes the algorithm faster. In \Cref{apxfig:runtime} [(B), (E)] we observe a similar trend as in \Cref{apxfig:runtime} [(A), (D)]. However, in this case Sinkorn algorithm with low regularization, $\varepsilon = 100$, has runtimes comparable to those of our method. Lastly, in \Cref{apxfig:runtime} [(C), (F)], we explicitly plot runtimes for Sinkhorn on both colored and grayscale images, for different values of the regularization parameter $\varepsilon$. In general, the algorithm on colored images is slower, and increasing the trimming threshold leads to higher runtimes. Moreover, we observe again that larger value of $\varepsilon$ makes the algorithms faster. 

\end{document}